\documentclass[10pt,twocolumn,letterpaper]{article}

\usepackage{cvpr}
\usepackage{times}
\usepackage{epsfig}
\usepackage{graphicx, subfig}
\usepackage{amsmath}
\usepackage{amssymb}
\usepackage{bm}
\usepackage{mathrsfs}
\usepackage{mathtools}
\usepackage{tabu}
\usepackage[font=small,labelfont=bf, belowskip=-3pt]{caption}
\usepackage{titling}

\newcommand{\cM}{\mathcal{M}}

\def \sH{{\mathscr{H}}}

\def \cA{{\mathcal{A}}}

\def \cD{{\mathcal{D}}}
\def \cF{{\mathcal{F}}}
\def \cH{{\mathcal{H}}}
\def \cJ{{\mathcal{J}}}
\def \cK{{\mathcal{K}}}
\def \cL{{\mathcal{L}}}
\def \cM{{\mathcal{M}}}
\def \cO{{\mathcal{O}}}
\def \cI{{\mathcal{I}}}
\def \cS{{\mathcal{S}}}

\def \cU{{\scriptstyle{\mathcal{U}}}}
\def \cV{{\scriptstyle{\mathcal{V}}}}

\def \bbE{{\mathbb{E}}}

\def \bbR{{\mathbb{R}}}

\def \bH{{\mathbf{H}}}

\def \bmh{{\bm{h}}}
\def \bmp{{\bm{p}}}

\def \bmu{{\bm{u}}}
\def \bmv{{\bm{v}}}
\def \bmw{{\bm{w}}}
\def \bmx{{\bm{x}}}
\def \bmy{{\bm{y}}}

\usepackage[pagebackref=true,breaklinks=true,letterpaper=true,colorlinks,bookmarks=false]{hyperref}

\cvprfinalcopy 

\def\cvprPaperID{2074} 

\ifcvprfinal\fi
\begin{document}

\title{Deep Hashing Network for Unsupervised Domain Adaptation}
\author{Hemanth Venkateswara, Jose Eusebio, Shayok Chakraborty, Sethuraman Panchanathan\\
Center for Cognitive Ubiquitous Computing, Arizona State University, Tempe, AZ, USA\\
{\tt\small \{hemanthv, jeusebio, shayok.chakraborty, panch\}@asu.edu}
}
\date{}
\maketitle

\begin{abstract}
In recent years, deep neural networks have emerged as a dominant machine learning tool for a wide variety of application domains. 
However, training a deep neural network requires a large amount of labeled data, which is an expensive process in terms of time, labor and human expertise. 
Domain adaptation or transfer learning algorithms address this challenge by leveraging labeled data in a different, but related source domain, to develop a model for the target domain. 
Further, the explosive growth of digital data has posed a fundamental challenge concerning its storage and retrieval. 
Due to its storage and retrieval efficiency, recent years have witnessed a wide application of hashing in a variety of computer vision applications. 
In this paper, we first introduce a new dataset, \textit{Office-Home}, to evaluate domain adaptation algorithms. 
The dataset contains images of a variety of everyday objects from multiple domains. 
We then propose a novel deep learning framework that can exploit labeled source data and unlabeled target data to learn informative hash codes, to accurately classify unseen target data. 
To the best of our knowledge, this is the first research effort to exploit the feature learning capabilities of deep neural networks to learn representative hash codes to address the domain adaptation problem. 
Our extensive empirical studies on multiple transfer tasks corroborate the usefulness of the framework in learning efficient hash codes which outperform existing competitive baselines for unsupervised domain adaptation. 
\end{abstract}

\section{Introduction}
Deep learning algorithms automatically learn a discriminating set of features and have depicted commendable performance in a variety of computer vision applications. 
Unfortunately, training a deep model necessitates a large volume of labeled data, which can be time consuming and expensive to acquire. 
However, labeled data from a different, but related domain is often available, which has motivated the development of algorithms which can leverage labeled data in a source domain to develop a machine learning model for the target domain. 
Learning a discriminative model in the presence of the shift between training and test distributions is known as transfer learning or domain adaptation \cite{ganin2016domain}. 
Unsupervised domain adaptation is a challenging setting, where labeled data is available only in the source domain; no labeled data is available in the target domain. 
Conventional shallow transfer learning methods develop their models in two stages, feature extraction followed by domain adaptation. 
The features are fixed and then a model is trained to align the source and target domains \cite{fernando2013unsupervised, gong2012geodesic, long2013transfer, pan2011domain, saenko2010adapting, shekhar2013generalized, sun2015return}. 
On the other hand, deep transfer learning procedures exploit the feature learning capabilities of deep networks to learn  transferable feature representations for domain adaptation and have demonstrated impressive empirical performance \cite{ganin2016domain, glorot2011domain, long2015learning, long2016unsupervised, tzeng2015simultaneous}. 

The explosive growth of digital data in the modern era has posed fundamental challenges regarding their storage, retrieval and computational requirements. 
Against this backdrop, hashing has emerged as one of the most popular and effective techniques due to its fast query speed and low memory cost \cite{wang2014hashing}. 
Hashing techniques transform high dimensional data into compact binary codes and generate similar binary codes for similar data items. 
Motivated by this fact, we propose to train a deep neural network to output binary hash codes (instead of probability values), which can be used for classification. 
We see two advantages to estimating a hash value instead of a standard probability vector in the final layer of the network: (i) the hash values are used to develop a unique loss function for target data in the absence of labels and (ii) during prediction, the hash value of a test sample can be compared against the hash values of the training samples to arrive at a more robust category prediction. 

In this paper, we first introduce a new dataset, \textit{Office-Home}, which we use to evaluate our algorithm. 
The \textit{Office-Home} dataset is an object recognition dataset which contains images from $4$ domains. 
It has around $15,500$ images organized into $65$ categories. 
We further propose a novel deep learning framework called Domain Adaptive Hashing (DAH) to learn informative hash codes to address the problem of unsupervised domain adaptation. 
We propose a unique loss function to train the deep network with the following components: (i) supervised hash loss for labeled source data, which ensures that source samples belonging to the same class have similar hash codes; (ii) unsupervised entropy loss for unlabeled target data, which imposes each target sample to align closely with exactly one of the source categories and be distinct from the other categories and (iii) a loss based on multi-kernel Maximum Mean Discrepancy (MK-MMD), which seeks to learn transferable features within the layers of the network to minimize the distribution difference between the source and target domains. 
Figure \ref{Fig:DAHnetwork} illustrates the different layers of the DAH and the components of the loss function. 

\section{Related Work}
There have been many approaches to address the problem of domain-shift in unsupervised domain adaptation. 
One straightforward approach is, to modify a classifier trained for the source data by adapting it to classify target data \cite{aytar2011tabula, bruzzone2010domain} or learn a transformation matrix to linearly transform the source data, so that it is aligned with the target \cite{hoffman2013, saenko2010adapting}. 
Some other procedures re-weight the data points in the source domain, to select source data that is similar to the target, when training a domain adaptive classifier, \cite{chattopadhyay2012multisource, chu2013selective, gong2013connecting}. 
A standard procedure to reduce domain discrepancy is, to project the source and target data to a common subspace, thereby aligning their principal axes \cite{fernando2013unsupervised, sun2015return}. 
Reducing domain disparity through nonlinear alignment of data has been possible with Maximum Mean Discrepancy (MMD) - a measure that provides the distribution difference between two datasets in a reproducing-kernel Hilbert space \cite{duan2012domain}.  
Kernel-PCA based methods apply the MMD to achieve nonlinear alignment of domains \cite{long2014transfer, long2013transfer, pan2011domain}. 
Manifold based approaches are also popular in domain adaptation for computer vision, where the subspace of a domain is treated as a point on the manifold and transformations are learned to align two domains \cite{gong2012geodesic, gopalan2011domain}. 
A survey of popular domain adaptation techniques for computer vision is provided in \cite{patel2015visual} and a more generic survey of transfer learning approaches can be found in \cite{pan2010survey}. 

All of the above techniques can be termed as shallow learning procedures, since the models are learned using pre-determined features. 
In recent years deep learning has become very successful at learning highly discriminative features for computer vision applications \cite{Chatfield14}. 
Deep learning systems like deep CNNs learn representations of data that capture underlying factors of variation between different tasks in a multi-task transfer learning setting \cite{bengio2013representation}. 
These representations also disentangle the factors of variation allowing for the transfer of knowledge between tasks \cite{donahue2014decaf, glorot2011domain, oquab2014learning}. 
Yosinski et al. \cite{yosinski2014transferable} demonstrated how the lower layers of a network produce generic features and the upper layers output task specific features. 
Based on this, deep learning procedures for domain adaptation train networks to learn transferable features in the fully connected final layers of a network \cite{long2015learning, tzeng2015simultaneous}. 
In other approaches to deep domain adaptation, Ganin et al. \cite{ganin2016domain} trained domain adversarial networks to learn features that make the source and target domain indistinguishable and Long et al. \cite{long2016unsupervised}, trained a network to do both feature adaptation and classifier adaptation using residual transfer networks. 

Unsupervised hashing techniques have been developed to extract unique hash codes for efficient storage and retrieval of data \cite{gong2013iterative, he2013k}. 
Neural network based hashing has led the way in state-of-the-art unsupervised hashing techniques \cite{carreira2015hashing, do2016learning, erin2015deep}. 
The closest work incorporating hashing and adaptation appears in cross-modal hashing, where deep hashing techniques embed multi-modal data and learn hash codes for two related domains, like text and images \cite{cao2016deep, cao2016transitive, jiang2016deep}. 
However, these algorithms are not unsupervised and they are mainly applied to extract common hash codes for multi-modal data for retrieval purposes. 
To the best of our knowledge, there has been no work in unsupervised domain adaptation using deep hashing networks. 
We now present the Domain Adaptive Hashing (DAH) network for unsupervised domain adaptation through deep hashing. 

\section{Domain Adaptive Hashing Networks}
In unsupervised domain adaptation, we consider data from two domains; \textit{source} and \textit{target}. 
The source consists of labeled data, $\cD_s = \{\bmx^s_i, y^s_i\}_{i=1}^{n_s}$ and the target has only unlabeled data $\cD_t = \{\bmx^t_i\}_{i=1}^{n_t}$. 
The data points $\bmx_i^*$ belong to $X$, where $X$ is some input space. The corresponding labels are represented by $y_i^* \in Y \coloneqq \{1,\ldots, C\}$. 
The paradigm of domain adaptive learning attempts to address the problem of \textit{domain-shift} in the data, where the data distributions of the source and target are different, i.e. $P_s(X,Y) \neq P_t(X,Y)$. 
The domain-shift notwithstanding, our goal is to train a deep neural network classifier $\psi(.)$, that can predict the labels $\{\hat{y}_i^t\}_{i=1}^{n_t}$, for the target data. 

We implement the neural network as a deep CNN which consists of 5 convolution layers \textit{conv}1 - \textit{conv}5 and 3 fully connected layers \textit{fc}6 - \textit{fc}8 followed by a loss layer. 
In our model, we introduce a hashing layer \textit{hash-fc}8 in place of the standard \textit{fc}8 layer to learn a binary code $\bmh_i$, for every data point $\bmx_i$, where $\bmh_i \in \{-1,+1\}^d$. 
The \textit{hash-fc}8 layer is driven by two loss functions, (i) \textit{supervised hash loss} for the source data, (ii) \textit{unsupervised entropy loss} for the target data. 
The supervised hash loss ensures hash values that are distinct and discriminatory, i.e. if $\bmx_i$ and $\bmx_j$ belong to the same category, their hash values $\bmh_i$ and $\bmh_j$ are similar and different otherwise. 
The unsupervised entropy loss aligns the target hash values with source hash values based on the similarity of their feature representations. 
The output of the network is represented as $\psi(\bmx)$, where $\psi(\bmx) \in \bbR^d$, which we convert to a hash code $\bmh = \text{sgn}(\psi(\bmx))$, where $\text{sgn}(.)$ is the sign function. 
Once the network has been trained, the probability of $\bmx$ being assigned a label $y$ is given by $f(\bmx) = p(y|\bmh)$. 
We train the network using $\cD_s$ and $\cD_t$ and predict the target data labels $\hat{y}_*^t$ using $f(.)$. 

In order to address the issue of domain-shift, we need to align the feature representations of the target and the source. 
We do that by reducing the domain discrepancy between the source and target feature representations at multiple layers of the network. 
In the following subsections, we discuss the design of the domain adaptive hash (DAH) network in detail. 
\begin{figure}[t]
\centering
		\includegraphics[width=\linewidth]{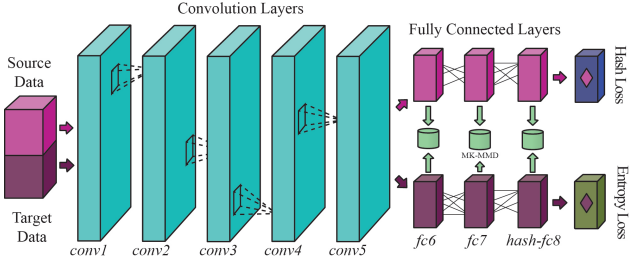}
\caption{The Domain Adaptive Hash (DAH) network that outputs hash codes for the source and the target. The network is trained with a batch of source and target data. The convolution layers \textit{conv}1 - \textit{conv}5  and the fully connected layers \textit{fc}6 and \textit{fc}7 are fine tuned from the VGG-F network. The MK-MMD loss trains the DAH to learn feature representations which align the source and the target. The \textit{hash-fc}8 layer is trained to output vectors of $d$ dimensions. The supervised hash loss drives the DAH to estimate a unique hash value for each object category. The unsupervised entropy loss aligns the target hash values to their corresponding source categories. Best viewed in color.}
\label{Fig:DAHnetwork}
\end{figure}

\subsection{Reducing Domain Disparity}
Deep learning methods have been very successful in domain adaptation with state-of-the-art algorithms \cite{ganin2016domain, long2015learning, long2016unsupervised, tzeng2015simultaneous} in recent years. 
The feature representations transition from generic to task-specific as one goes up the layers of a deep CNN \cite{yosinski2014transferable}. 
The convolution layers \textit{conv}1 to \textit{conv}5 have been shown to be generic and so, readily transferable, whereas the fully connected layers are more task-specific and need to be adapted before they can be transferred. 
In the DAH algorithm, we attempt to minimize the MK-MMD loss to reduce the domain difference between the source and target feature representations for fully connected layers, $\cF = \{\textit{fc}6, \textit{fc}7, \textit{fc}8\}$. 
Such a loss function has been used in previous research \cite{long2015learning, long2016unsupervised}. 
The multi-layer MK-MMD loss is given by, 
\begin{flalign}
\cM(\cU_s,\cU_t) = \sum_{l\in\cF}d_k^2(\cU^l_s,\cU^l_t),
\label{Eq:MKMMD}
\end{flalign}
where, $\cU^l_s = \{\bmu_i^{s,l}\}_{i=1}^{n_s}$ and $\cU^l_t = \{\bmu_i^{t,l}\}_{i=1}^{n_t}$ are the set of output representations for the source and target data at layer $l$, where $\bmu_i^{*,l}$ is the output representation of $\bmx_i^*$ for the $l^{th}$ layer. 
The final layer outputs are denoted as $\cU_s$ and $\cU_t$. 
The MK-MMD measure $d_k^2(.)$ is the multi-kernel maximum mean discrepancy between the source and target representations, \cite{gretton2012optimal}. 
For a nonlinear mapping $\phi(.)$ associated with a reproducing kernel Hilbert space $\sH_k$ and kernel $k(.)$, where $k(\bmx,\bmy) = \langle\phi(\bmx),\phi(\bmy)\rangle$, the MMD is defined as,
\begin{flalign}
d_k^2(\cU^l_s,\cU^l_t)=\Big|\Big|\bbE[\phi(\bmu^{s,l})] - \bbE[\phi(\bmu^{t,l})]\Big|\Big|_{\sH_k}^2.
\label{Eq:layerMKMMD}
\end{flalign}
The characteristic kernel $k(.)$, is determined as a convex combination of $\kappa$ PSD kernels, $\{k_m\}_{m=1}^\kappa$, 
$\cK\coloneqq\big\{k:k=\sum_{m=1}^\kappa\beta_m k_m, \sum_{m=1}^\kappa\beta_m=1, \beta_m\geq 0, \forall m\big\}$. 
We set $\beta_m = 1/\kappa$ according to \cite{long2016unsupervised} and it works well in practice. 

\subsection{Supervised Hashing for Source Data}
The Hamming distance for a pair of hash values $\bmh_i$ and $\bmh_j$ has a unique relationship with the dot product $\langle\bmh_i, \bmh_j\rangle$, given by: $\text{dist}_H (\bmh_i, \bmh_j) = \frac{1}{2}(d - \bmh_i^\top \bmh_j)$, where $d$ is the hash length.  
The dot product $\langle\bmh_i, \bmh_j\rangle$ can be treated as a similarity measure for the hash codes. 
Larger the value of the dot product (high similarity), smaller is the distance $\text{dist}_H$ and smaller the dot product (low similarity), larger is the distance $\text{dist}_H$. 
Let $s_{ij} \in \{0, 1\}$ be the similarity between $\bmx_i$ and $\bmx_j$. 
If $\bmx_i$ and $\bmx_j$ belong to the same category, $s_{ij} = 1$ and 0, otherwise. 
The probability of similarity between $\bmx_i$ and $\bmx_j$ given the corresponding hash values $\bmh_i$ and $\bmh_j$, can be expressed as a likelihood function, given by,
\begin{flalign}
p(s_{ij}|\bmh_i,\bmh_j) = 
	\begin{cases}
	\sigma(\bmh_i^\top \bmh_j), &\quad s_{ij} = 1\\
	1-\sigma(\bmh_i^\top \bmh_j), &\quad s_{ij} = 0,
	\end{cases}
	\label{Eq:hashProb}
\end{flalign}
where, $\sigma(x) = \frac{1}{1+e^{-x}}$ is the sigmoid function. 
As the dot product $\langle\bmh_i, \bmh_j\rangle$ increases, the probability of $p(s_{ij}=1|\bmh_i,\bmh_j)$ also increases, i.e., $\bmx_i$ and $\bmx_j$ belong to the same category. 
As the dot product decreases, the probability $p(s_{ij}=1|\bmh_i,\bmh_j)$ also decreases, i.e., $\bmx_i$ and $\bmx_j$ belong to different categories. 
We construct the $(n_s \times n_s)$ similarity matrix $\cS = \{s_{ij}\}$, for the source data with the provided labels, where $s_{ij} = 1$ if $\bmx_i$ and $\bmx_j$ belong to the same category and 0, otherwise. 
Let $\bH = \{\bmh_i\}_{i=1}^{n_s}$ be the set of source data hash values. 
If the elements of $\bH$ are assumed to be i.i.d., the negative log likelihood of the similarity matrix $\cS$ given $\bH$ can be written as,
\begin{flalign}
\min_{\bH} \cL(\bH) &= -\text{log}~p(\cS|\bH) \notag \\
		&= -\sum_{s_{ij} \in \cS}\Big( s_{ij}\bmh_i^\top \bmh_j - \text{log}\big(1 + \text{exp}(\bmh_i^\top \bmh_j)\big)\Big).
\label{Eq:loglikelihood}
\end{flalign}
By minimizing Equation (\ref{Eq:loglikelihood}), we can determine hash values $\bH$ for the source data which are consistent with the similarity matrix $\cS$. 
The hash loss has been used in previous research for supervised hashing \cite{li2015feature, zhu2016deep}. 
Equation (\ref{Eq:loglikelihood}) is a discrete optimization problem that is challenging to solve. 
We introduce a relaxation on the discrete constraint $\bmh_i \in \{-1,+1\}^d$ by instead solving for $\bmu_i \in \bbR^d$, where $\cU_s = \{\bmu_i\}_{i=1}^{n_s}$ is the output of the network and $\bmu_i = \psi(\bmx_i)$ (the superscript denoting the domain has been dropped for ease of representation). 
However, the continuous relaxation gives rise to (i) approximation error, when $\langle\bmh_i, \bmh_j\rangle$ is substituted with $\langle\bmu_i, \bmu_j\rangle$ and, (ii) quantization error, when the resulting real codes $\bmu_i$ are binarized \cite{zhu2016deep}. 
We account for the approximation error by having a $\text{tanh}(.)$ as the final activation layer of the neural network, so that the components of $\bmu_i$ are bounded between $-1$ and $+1$. 
In addition, we also introduce a quantization loss $||\bmu_i - \text{sgn}(\bmu_i)||_2^2$ along the lines of \cite{gong2013iterative}, where $\text{sgn}(.)$ is the sign function. 
The continuous optimization problem for supervised hashing can now be outlined; 
\begin{flalign}
\min_{\cU_s} \cL(\cU_s) =& -\sum_{s_{ij} \in \cS}\Big( s_{ij}\bmu_i^\top \bmu_j - \text{log}\big(1 + \text{exp}(\bmu_i^\top \bmu_j)\big)\Big)\notag\\
		& + \sum_{i=1}^{n_s}\big|\big|\bmu_i - \text{sgn}(\bmu_i)\big|\big|_2^2.
\label{Eq:supHashLoss}
\end{flalign}

\subsection{Unsupervised Hashing for Target Data}
In the absence of target data labels, we use the similarity measure $\langle\bmu_i, \bmu_j\rangle$, to guide the network to learn discriminative hash values for the target data. 
An ideal target output $\bmu_i^t$, needs to be similar to many of the source outputs from the $j^{th}$ category $\big(\{\bmu_k^{s_j}\}_{k=1}^K\big)$. 
We assume without loss of generality, $K$ source data points for every category $j$ where, $j \in \{1, \ldots, C\}$ and $\bmu_k^{s_j}$ is the $k^{th}$ source output from category $j$. 
In addition, $\bmu_i^t$ must be dissimilar to most other source outputs $\bmu_k^{s_l}$ belonging to a different category ($j\neq l$). 
Enforcing similarity with all the $K$ data points makes for a more robust target data category assignment. 
We outline a probability measure to capture this intuition. 
Let $p_{ij}$ be the probability that input target data point $\bmx_i$ is assigned to category $j$ where,
\begin{flalign}
p_{ij} = \frac{\sum_{k=1}^K \text{exp}({\bmu_i^t}^\top \bmu_k^{s_j})}{\sum_{l=1}^C\sum_{k=1}^K \text{exp}({\bmu_i^t}^\top \bmu_k^{s_l})}
\label{Eq:EntProb}
\end{flalign}
The $\text{exp}(.)$ has been introduced for ease of differentiability and the denominator ensures $\sum_jp_{ij} = 1$. 
When the target data point output is similar to one category only and dissimilar to all the other categories, the probability vector $\bmp_i = [p_{i1},\ldots,p_{iC}]^T$ tends to be a one-hot vector. 
A one-hot vector can be viewed as a low entropy realization of $\bmp_i$. 
We can therefore envisage all the $\bmp_i$ to be one-hot vectors (low entropy probability vectors), where the target data point outputs are similar to source data point outputs in one and only one category. 
To this end we introduce a loss to capture the entropy of the target probability vectors. 
The entropy loss for the network outputs is given by, 
\begin{flalign}
\cH(\cU_s, \cU_t) = -\frac{1}{n_t}\sum_{i = 1}^{n_t}\sum_{j=1}^C p_{ij}\text{log}(p_{ij})
\label{Eq:EntLoss}
\end{flalign}
\noindent Minimizing the entropy loss gives us probability vectors $\bmp_i$ that tend to be one-hot vectors, i.e., the target data point outputs are similar to source data point outputs from any one category only. 
Enforcing similarity with $K$ source data points from a category, guarantees that the hash values are determined based on a common similarity between multiple source category data points and the target data point. 

\subsection{Domain Adaptive Hash Network}
We propose a model for deep unsupervised domain adaptation based on hashing (DAH) that incorporates unsupervised domain adaptation between the source and the target (\ref{Eq:MKMMD}), supervised hashing for the source (\ref{Eq:supHashLoss}) and unsupervised hashing for the target (\ref{Eq:EntLoss}) in a deep convolutional neural network. The DAH network is trained to minimize
\begin{flalign}
\min_{\cU}\cJ = \cL(\cU_s) + \gamma\cM(\cU_s,\cU_t) + \eta\cH(\cU_s, \cU_t),
\label{Eq:DAH}
\end{flalign}
where, $\cU\coloneqq\{\cU_s\cup~\cU_t\}$ and ($\gamma$, $\eta$) control the importance of domain adaptation (\ref{Eq:MKMMD}) and target entropy loss (\ref{Eq:EntLoss}) respectively. 
The hash values $\bH$ are obtained from the output of the network using $\bH = \text{sgn}(\cU)$. 
The loss terms (\ref{Eq:supHashLoss}) and (\ref{Eq:EntLoss}) are determined in the final layer of the network with the network output $\cU$. 
The MK-MMD loss (\ref{Eq:MKMMD}) is determined between layer outputs $\{\cU_s^l, \cU_t^l\}$ at each of the fully connected layers $\cF =
 \{\textit{fc}6, \textit{fc}7, \textit{fc}8\}$, 
where we adopt the linear time estimate for the unbiased MK-MMD as described in \cite{gretton2012optimal} and \cite{long2015learning}. 
The DAH is trained using standard back-propagation. 
The detailed derivation of the derivative of (\ref{Eq:DAH}) w.r.t. $\cU$ is provided in the supplementary material. 

\noindent\textbf{Network Architecture}: Owing to the paucity of images in a domain adaptation setting, we circumvent the need to train a deep CNN with millions of images by adapting the pre-trained VGG-F \cite{Chatfield14} network to the DAH. 
The VGG-F has been trained on the ImageNet 2012 dataset and it consists of 5 convolution layers (\textit{conv}1 - \textit{conv}5) and 3 fully connected layers (\textit{fc}6, \textit{fc}7, \textit{fc}8). 
We introduce the hashing layer \textit{hash-fc}8 that outputs vectors in $\bbR^d$ in the place of \textit{fc}8. 
To account for the hashing approximation, we introduced a $\text{tanh}()$ layer. 
However, we encounter the issue of vanishing gradients \cite{hochreiter2001gradient} when using $\text{tanh}()$ as it saturates with large inputs. 
We therefore preface the $\text{tanh}()$ with a batch normalization layer which prevents the $\text{tanh}()$ from saturating. 
In effect, $\textit{hash-fc}8 \coloneqq \{\textit{fc}8 \rightarrow \textit{batch-norm} \rightarrow \textit{tanh}()\}$. 
The $\textit{hash-fc}8$ provides greater stability when fine-tuning the learning rates than the deep hashing networks \cite{li2015feature, zhu2016deep}. 
Figure \ref{Fig:DAHnetwork} illustrates the proposed DAH network. 

\section{The \textit{Office-Home} Dataset}
\begin{figure*}[t]
\centering
		\includegraphics[width=\textwidth]{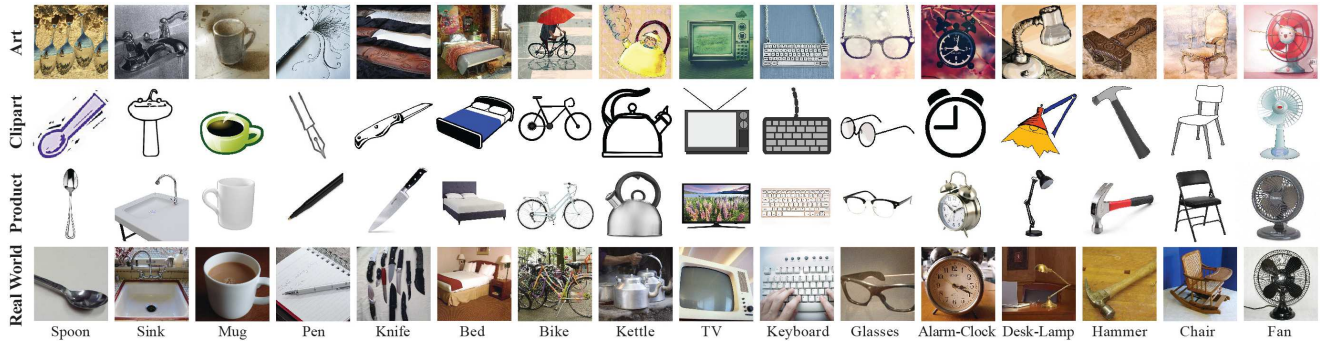}
\caption{Sample images from the \textbf{\textit{Office-Home}} dataset. The dataset consists of images of everyday objects organized into 4 domains; \texttt{Art}: paintings, sketches and/or artistic depictions, \texttt{Clipart}: clipart images, \texttt{Product}: images without background and \texttt{Real-World}: regular images captured with a camera. The figure displays examples from 16 of the 65 categories.}
\label{Fig:OfficeHomeDataset}
\end{figure*}
Supervised deep learning models require a large volume of labeled training data. 
Unfortunately, existing datasets for vision-based domain adaptation are limited in their size and are not suitable for validating deep learning algorithms. 
The standard datasets for vision based domain adaptation are, facial expression datasets \textit{CKPlus} \cite{lucey2010extended} and \textit{MMI} \cite{pantic2005web}, digit datasets \textit{SVHN} \cite{37648}, \textit{USPS} and \textit{MNIST}\cite{jarrett2009best}, head pose recognition datasets \textit{PIE} \cite{long2013transfer}, object recognition datasets \textit{COIL}\cite{long2013transfer}, \textit{Office} \cite{saenko2010adapting} and \textit{Office-Caltech} \cite{gong2012geodesic}. 
These datasets were created before deep-learning became popular and are insufficient for training and evaluating deep learning based domain adaptation approaches. 
For instance, the object-recognition dataset \textit{Office} has $4110$ images across $31$ categories and \textit{Office-Caltech} has $2533$ images across $10$ categories. 

We release the \textbf{\textit{Office-Home}} dataset for domain adaptation based object recognition, that can be used to evaluate deep learning algorithms for domain adaptation. 
The \textit{Office-Home} dataset consists of $4$ domains, with each domain containing images from $65$ categories of everyday objects and a total of around $15,500$ images. The domains include, 
\noindent\texttt{Art}: artistic depictions of objects in the form of sketches, paintings, ornamentation, etc.; 
\noindent\texttt{Clipart}: collection of clipart images; 
\noindent\texttt{Product}: images of objects without a background, akin to the Amazon category in \textit{Office} dataset; 
\noindent\texttt{Real-World}: images of objects captured with a regular camera. 
\begin{table}[t]
\centering
\caption{Statistics for the \textit{Office-Home} dataset. \textbf{Min: \#} is the minimum number of images amongst all the categories, \textbf{Min: Size} and \textbf{Max: Size} are the minimum and maximum image sizes across all categories and \textbf{Acc.} is the classification accuracy.}
\label{Tab:OfficeHomeStats}
\resizebox{\linewidth}{!}{%
\begin{tabu}{|c|[2pt] c c c c|}
\hline
 \textbf{Domain.} &  \textbf{Min: \#} &  \textbf{Min: Size} &  \textbf{Max: Size} & \textbf{Acc} \\\tabucline[2pt]{-}
 \texttt{Art}  & 15  & 117$\times$85 pix.  & 4384$\times$2686 pix.  & 44.99$\pm$1.85 \\
 \texttt{Clipart}  & 39  & 18$\times$18 pix.  & 2400$\times$2400 pix.  & 53.95$\pm$1.45 \\
 \texttt{Product}  & 38  & 75$\times$63 pix. & 2560$\times$2560 pix.  & 66.41$\pm$1.18 \\
 \texttt{Real-World}  & 23  & 88$\times$80 pix.  & 6500$\times$4900 pix.  & 59.70$\pm$1.04 \\\hline
\end{tabu}
}
\end{table}

Public domain images were downloaded from websites like www.deviantart.com and www.flickr.com to create the \texttt{Art} and \texttt{Real-World} domains. 
\texttt{Clipart} images were gathered from multiple clipart websites. 
The \texttt{Product} domain images were exclusively collected from www.amazon.com using web-crawlers. 
The collected images were manually filtered on the basis of quality, size and content. 
The dataset has an average of around $70$ images per category and a maximum of $99$ images in a category. 
The primary challenge in creating this dataset was acquiring sufficient number of public domain images across all the $4$ domains. 
Figure \ref{Fig:OfficeHomeDataset} depicts a sampling of $16$ categories from the \textit{Office-Home} dataset and Table \ref{Tab:OfficeHomeStats} outlines some meta data for the dataset. 
The \textbf{Acc.} column in the Table \ref{Tab:OfficeHomeStats} refers to classification accuracies using the LIBLINEAR SVM \cite{fan2008liblinear} classifier (5-fold cross validation) with deep features extracted using the VGG-F network. 
The dataset is publicly available for research \footnote{\texttt{https://hemanthdv.github.io/officehome-dataset/}}. 

\section{Experiments}
In this section we conduct extensive experiments to evaluate the DAH algorithm. 
Since we propose a domain adaptation technique based on hashing, we evaluate objection recognition accuracies for unsupervised domain adaptation and also study the discriminatory capability of the learned hash codes for unsupervised domain adaptive hashing. 
The implementation details are available at {\small\texttt{https://github.com/hemanthdv/da-hash}}

\subsection{Datasets}
\noindent\textbf{\textit{Office}} \cite{saenko2010adapting}: This is currently the most popular benchmark dataset for object recognition in the domain adaptation computer vision community. 
The dataset consists of images of everyday objects in an office environment. 
It has $3$ domains; \texttt{Amazon} (\textbf{A}), \texttt{Dslr} (\textbf{D}) and \texttt{Webcam} (\textbf{W}). 
The dataset has around $4,100$ images with a majority of the images ($2816$ images) in the \texttt{Amazon} domain. 
We adopt the common evaluation protocol of different pairs of transfer tasks for this dataset \cite{long2015learning, long2016unsupervised}. 
We consider 6 transfer tasks for all combinations of source and target pairs for the $3$ domains. 

\noindent\textbf{\textit{Office-Home}}: We introduce this new dataset and evaluate it in a similar manner to the \textit{Office} dataset. 
We consider 12 transfer tasks for the \texttt{Art} (\textbf{Ar}), \texttt{Clipart} (\textbf{Cl}), \texttt{Product} (\textbf{Pr}) and \texttt{Real-World} (\textbf{Rw}) domains for all combinations of source and target for the $4$ domains. 
Considering all the different pairs of transfer enables us to evaluate the inherent bias between the domains in a comprehensive manner \cite{torralba2011unbiased}. 

\subsection{Implementation Details}
We implement the DAH using the MatConvnet framework \cite{vedaldi15matconvnet}. 
Since we train a pre-trained VGG-F, we fine-tune the weights of \textit{conv}1-\textit{conv}5, \textit{fc}6 and \textit{fc}7. 
We set their learning rates to ${1/10}^{th}$ the learning rate of \textit{hash-fc}8. 
We vary the learning rate between $10^{-4}$ to $10^{-5}$ over $300$ epochs with a momentum $0.9$ and weight decay $5\times10^{-4}$. 
We set $K=5$ (number of samples from a category). 
Since we have $31$ categories in the \textit{Office} dataset, we get a source batch size of $31\times5 = 155$. 
For the target batch, we randomly select $155$ samples. 
The total batch size turns out to be $310$. 
For the \textit{Office-Home} dataset, with $K=5$ and $65$ categories, we get a batch size of $650$. 
We set $d=64$ (hash code length) for all our experiments. 
Since there is imbalance in the number of like and unlike pairs in $\cS$, we set the values in similarity matrix $\cS_{i,j} \in \{0,10\}$. Increasing the similarity weight of like-pairs improves the performance of DAH. 
For the entropy loss, we set $\eta = 1$. 
For the MK-MMD loss, we follow the heuristics mentioned in \cite{gretton2012optimal}, to determine the parameters. 
We estimate $\gamma$, by validating a binary domain classifier to distinguish between source and target data points and select $\gamma$ which gives largest error on a validation set. 
For MMD, we use a Gaussian kernel with a bandwidth $\sigma$ given by the median of the pairwise distances in the training data. 
To incorporate the multi-kernel, we vary the bandwidth $\sigma_m\in[2^{-8}\sigma, 2^{8}\sigma]$ with a multiplicative factor of 2. 
We define the target classifier $f(\bmx^t_i) = p(y|\bmh^t_i)$ in terms of \ref{Eq:EntProb}. 
The target data point is assigned to the class with the largest probability, with $\hat{y}_i = \max_j(p_{ij})$ using the hash codes for the source and the target. 

\subsection{Unsupervised Domain Adaptation}
In this section, we study the performance of the DAH for unsupervised domain adaptation, where labeled data is available only in the source domain and no labeled data is available in the target domain. 
We compare the DAH with state-of-the-art domain adaptation methods: (i) Geodesic Flow Kernel (\textbf{GFK}) \cite{gong2012geodesic}, (ii) Transfer Component Analysis (\textbf{TCA}) \cite{pan2011domain}, (iii) Correlation Alignment (\textbf{CORAL}) \cite{sun2015return} and (iv) Joint Distribution Adaptation (\textbf{JDA}) \cite{long2013transfer}. 
We also compare the DAH with state-of-the-art deep learning methods for domain adaptation: (v) Deep Adaptation Network (\textbf{DAN}) \cite{long2015learning} and (vi) Domain Adversarial Neural Network (\textbf{DANN}) \cite{ganin2016domain}.  
For all of the shallow learning methods, we extract and use deep features from the \textit{fc}7 layer of the VGG-F network that was pre-trained on the ImageNet 2012 dataset. 
We also evaluate the effect of the entropy loss on hashing for the DAH. 
The \textbf{DAH-e} is the DAH algorithm where $\eta$ is set to zero, which implies that the target hash values are not driven to align with the source categories. 
We follow the standard protocol for unsupervised domain adaptation, where all the labeled source data and all the unlabeled target data is used for training. 

\noindent\textbf{Results and Discussion}: The results are reported for the target classification in each of the transfer tasks in Tables \ref{Tab:OfficeAccDAH} and \ref{Tab:OfficeHomeAccDAH}, where accuracies denote the percentage of correctly classified target data samples. 
We present results with hash length $d=64$ bits. 
The DAH algorithm consistently outperforms the baselines across all the domains for the \textit{Office-Home} dataset. 
However, DANN marginally surpasses DAH for the \textit{Office} dataset, prompting us to reason that domain adversarial training is more effective than DAH when the categories are fewer in number. 
Since domain alignment is category agnostic, it is possible that the aligned domains are not classification friendly in the presence of large number of categories. 
When the number of categories is large, as in \textit{Office-Home}, DAH does best at extracting transferable features to achieve higher accuracies. 
We also note that DAH delivers better performance than DAH-e; thus, minimizing the entropy on the target data through \ref{Eq:EntLoss} aids in improved alignment of the source and target samples, which boosts the accuracy. 

\begin{table}[t]
\centering
\caption{Recognition accuracies (\%) for domain adaptation experiments on the \textit{Office} dataset. \{\texttt{Amazon} (A), \texttt{Dslr} (D), \texttt{Webcam} (W)\}. A$\rightarrow$W implies A is source and W is target.}
\label{Tab:OfficeAccDAH}
\resizebox{\linewidth}{!}{%
\begin{tabu}{|c |[2pt] c c c c c c | c|}
\hline
\textbf{Expt.} &  \textbf{A$\rightarrow$D} &  \textbf{A$\rightarrow$W} &  \textbf{D$\rightarrow$A} &  \textbf{D$\rightarrow$W} &  \textbf{W$\rightarrow$A} &  \textbf{W$\rightarrow$D} & \textbf{Avg.} \\\tabucline[2pt]{-}
 GFK  & 48.59  & 52.08  & 41.83  & 89.18  & 49.04  & 93.17  & 62.32 \\
 TCA  & 51.00  & 49.43  & 48.12  & 93.08  & 48.83  & 96.79  & 64.54 \\
 CORAL  & 54.42  & 51.70  & 48.26  & 95.97  & 47.27  & 98.59  & 66.04 \\
 JDA  & 59.24  & 58.62  & 51.35  & 96.86  & 52.34  & 97.79  & 69.37 \\
 DAN  & 67.04  & 67.80  & 50.36  & 95.85  & 52.33  & 99.40  & 72.13 \\
 DANN  & 72.89  & 72.70  & 56.25  & 96.48  & 53.20  & 99.40  & 75.15 \\\hline
 DAH-e  & 66.27  & 66.16  & 55.97  & 94.59  & 53.91  & 96.99  & 72.31 \\
 DAH  & 66.47  & 68.30  & 55.54  & 96.10  & 53.02  & 98.80  & 73.04 \\\hline
\end{tabu}
}
\end{table}

\begin{table*}[t]
\centering
\caption{Recognition accuracies (\%) for domain adaptation experiments on the \textit{Office-Home} dataset. \{\texttt{Art} (Ar), \texttt{Clipart} (Cl), \texttt{Product} (Pr), \texttt{Real-World} (Rw)\}. Ar$\rightarrow$Cl implies Ar is source and Cl is target.}
\label{Tab:OfficeHomeAccDAH}
\resizebox{\textwidth}{!}{%
\begin{tabu}{c |[2pt] c c c c c c c c c c c c | c}
\hline
\textbf{Expt.} &  \textbf{Ar$\rightarrow$Cl} &  \textbf{Ar$\rightarrow$Pr} &  \textbf{Ar$\rightarrow$Rw} &  \textbf{Cl$\rightarrow$Ar} &  \textbf{Cl$\rightarrow$Pr} &  \textbf{Cl$\rightarrow$Rw} &  \textbf{Pr$\rightarrow$Ar} &  \textbf{Pr$\rightarrow$Cl} &  \textbf{Pr$\rightarrow$Rw} &  \textbf{Rw$\rightarrow$Ar} &  \textbf{Rw$\rightarrow$Cl} &  \textbf{Rw$\rightarrow$Pr} & \textbf{Avg.} \\\tabucline[2pt]{-}
 GFK  & 21.60  & 31.72  & 38.83  & 21.63  & 34.94  & 34.20  & 24.52  & 25.73  & 42.92  & 32.88  & 28.96  & 50.89  & 32.40 \\
 TCA  & 19.93  & 32.08  & 35.71  & 19.00  & 31.36  & 31.74  & 21.92  & 23.64  & 42.12  & 30.74  & 27.15  & 48.68  & 30.34 \\
 CORAL  & 27.10  & 36.16  & 44.32  & 26.08  & 40.03  & 40.33  & 27.77  & 30.54  & 50.61  & 38.48  & 36.36  & 57.11  & 37.91 \\
 JDA  & 25.34  & 35.98  & 42.94  & 24.52  & 40.19  & 40.90  & 25.96  & 32.72  & 49.25  & 35.10  & 35.35  & 55.35  & 36.97 \\
 DAN  & 30.66  & 42.17  & 54.13  & 32.83  & 47.59  & 49.78  & 29.07  & 34.05  & 56.70  & 43.58  & 38.25  & 62.73  & 43.46 \\
 DANN  & 33.33  & 42.96  & 54.42  & 32.26  & 49.13  & 49.76  & 30.49  & 38.14  & 56.76  & 44.71  & 42.66  & 64.65  & 44.94 \\\hline
 DAH-e  & 29.23  & 35.71  & 48.29  & 33.79  & 48.23  & 47.49  & 29.87  & 38.76  & 55.63  & 41.16  & 44.99  & 59.07  & 42.69 \\
 DAH  & 31.64  & 40.75  & 51.73  & 34.69  & 51.93  & 52.79  & 29.91  & 39.63  & 60.71  & 44.99  & 45.13  & 62.54  & 45.54 \\\hline
\end{tabu}
}
\end{table*}

\begin{figure*}[t]
\centering
\subfloat[\scriptsize{$\cA$-Distance}]{
		\label{Fig:distA}
    \includegraphics[trim = 3mm 5mm 10mm 9mm, clip, width=0.245\textwidth]{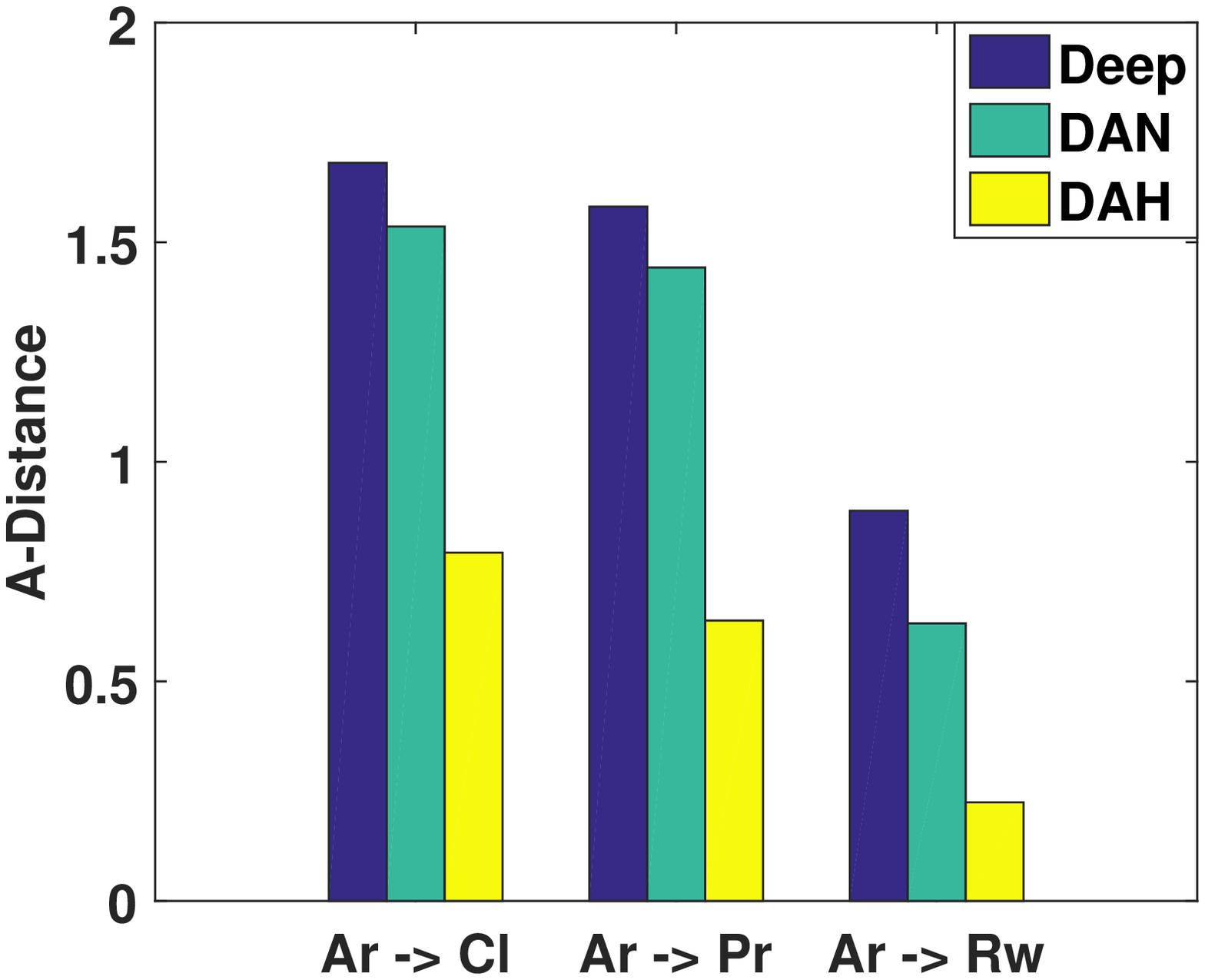}
}%
\hfill
\subfloat[\scriptsize{Deep Features (\textbf{Ar},\textbf{Cl})}]{
		\label{Fig:DeeptSNEArCl}
    \includegraphics[trim = 26mm 18mm 25mm 15mm, clip, width=0.2\textwidth]{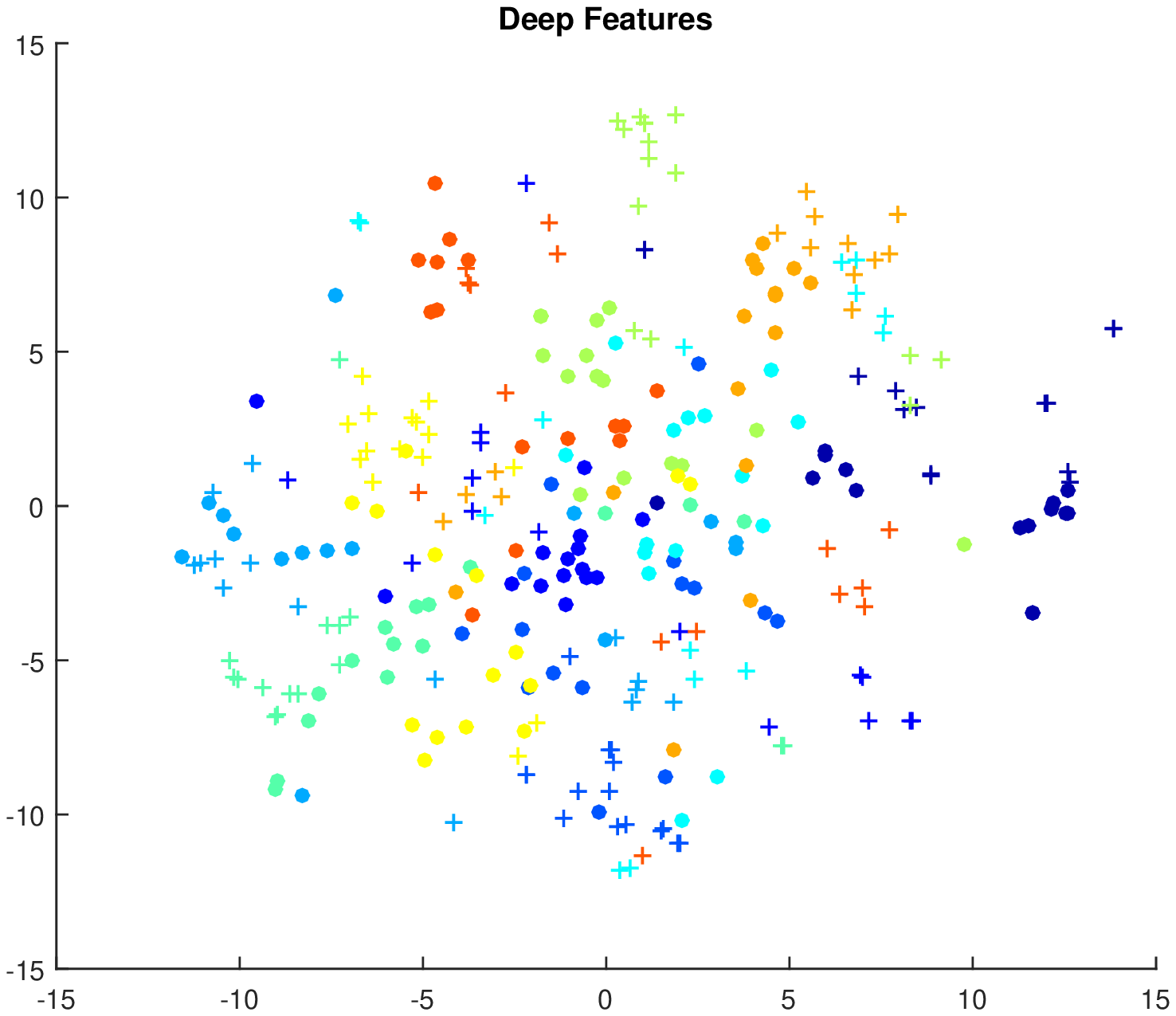}
}%
\hfill
\subfloat[\scriptsize{DAN Features (\textbf{Ar},\textbf{Cl})}]{
		\label{Fig:DANtSNEArCl}
    \includegraphics[trim = 26mm 18mm 25mm 15mm, clip, width=0.2\textwidth]{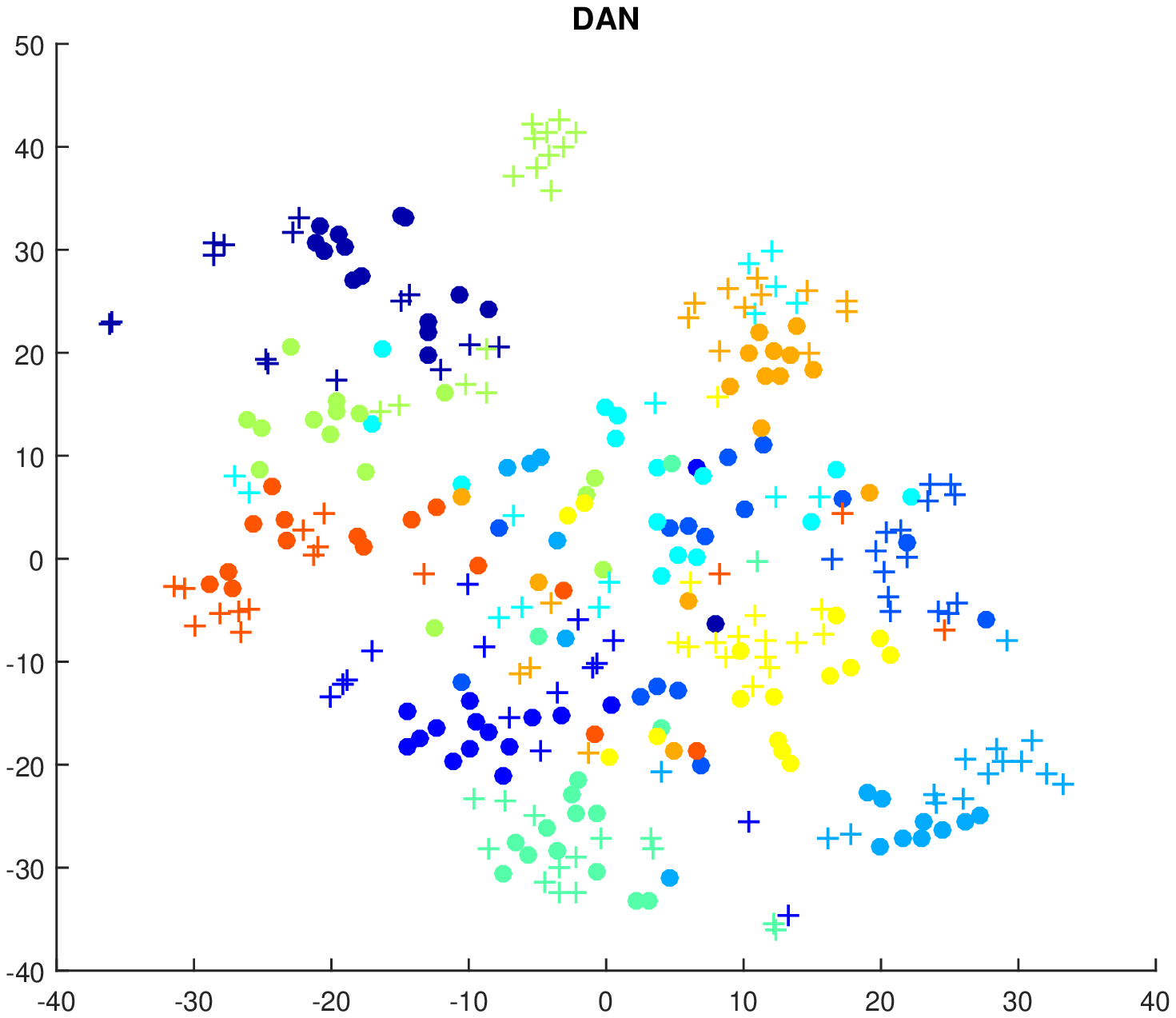}
}%
\hfill
\subfloat[\scriptsize{DAH Features (\textbf{Ar},\textbf{Cl})}]{
		\label{Fig:DAHtSNEArCl}
    \includegraphics[trim = 26mm 18mm 25mm 15mm, clip, width=0.2\textwidth]{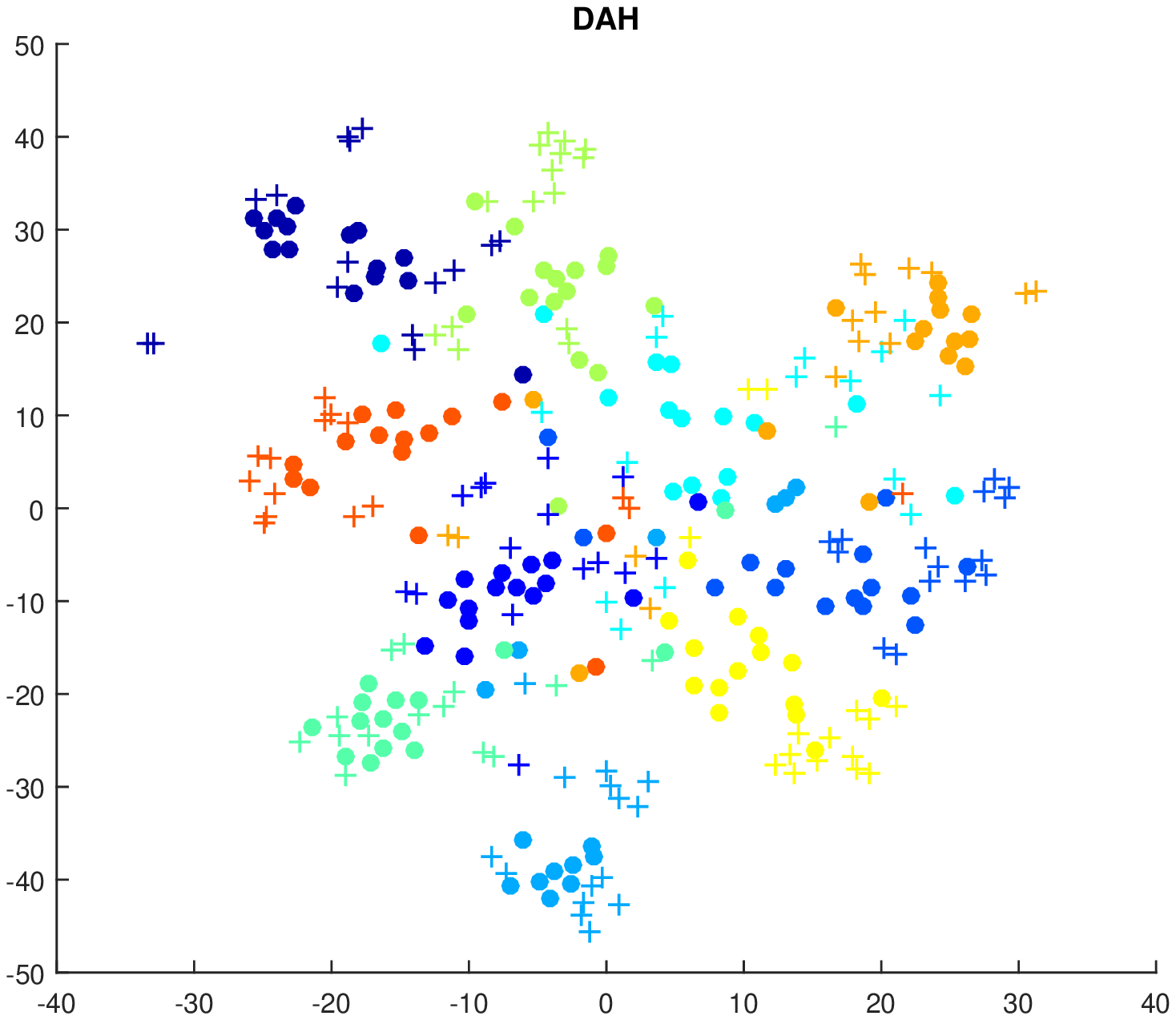}
}%
\caption{Feature analysis of \textit{fc}7 layer. (a) $\cA$-distances for Deep, DAN and DAH, (b), (c) and (d) t-SNE embeddings for 10 categories from \texttt{Art} ($\bullet$) and \texttt{Clipart}(+) domains. Best viewed in color.}
\label{Fig:dist-tSNE-ArCl}
\end{figure*}

\noindent\textbf{Feature Analysis}: We also study the feature representations of the penultimate layer (\textit{fc}7) outputs using t-SNE embeddings as in \cite{donahue2014decaf}. 
Figure \ref{Fig:distA} depicts the $\cA$-distance between domain pairs using Deep (VGG-F), DAN and DAH features. 
Ben-David et al. \cite{ben2010theory} defined $\cA$-distance as the distance between two domains that can be viewed as the discrepancy between two domains. 
Although it is difficult to estimate its exact value, an approximate distance measure is given by $2(1-2\epsilon)$, where $\epsilon$ is the generalization error for a binary classifier trained to distinguish between the two domains. 
We used a LIBLINEAR SVM \cite{fan2008liblinear} classifier with 5-fold cross-validation to estimate $\epsilon$. 
Figure \ref{Fig:distA} indicates that the DAH features have the least discrepancy between the source and target compared to DAN and Deep features. 
This is also confirmed with the t-SNE embeddings in Figures \ref{Fig:DeeptSNEArCl}-\ref{Fig:DAHtSNEArCl}. 
The Deep features show very little overlap between the domains and the categories depict minimal clustering. 
Domain overlap and clustering improves as we move to DAN and DAH features, with DAH providing the best visualizations. 
This corroborates the efficacy of the DAH algorithm to exploit the feature learning capabilities of deep neural networks to learn representative hash codes to address domain adaptation. 

\subsection{Unsupervised Domain Adaptive Hashing}
In this section, we study the performance of our algorithm to generate compact and efficient hash codes from the data for classifying unseen test instances, when no labels are available. 
This problem has been addressed in the literature, with promising empirical results \cite{carreira2015hashing, do2016learning, gong2011iterative}. 
However, in a real-world setting, labels may be available from a different, but related (source) domain; a strategy to utilize the labeled data from the source domain to learn representative hash codes for the target domain is therefore of immense practical importance. 
Our work is the first to identify and address this problem. 
We consider the following scenarios to address this real-world challenge: (i) No labels are available for a given dataset and the hash codes need to be learned in a completely unsupervised manner.
We evaluate against baseline unsupervised hashing methods (\textbf{ITQ}) \cite{gong2013iterative} and (\textbf{KMeans}) \cite{he2013k} and also state-of-the-art methods for unsupervised hashing (\textbf{BA}) \cite{carreira2015hashing} and (\textbf{BDNN}) \cite{do2016learning}. 
(ii) Labeled data is available from a different, but related source domain. 
A hashing model is trained on the labeled source data and is used to learn hash codes for the target data. 
We refer to this method as \textbf{NoDA}, as no domain adaptation is performed. 
We used the deep pairwise-supervised hashing (DPSH) algorithm \cite{li2015feature} to train a deep network with the source data and applied the network to generate hash codes for the target data. 
(iii) Labeled data is available from a different, but related source domain and we use our DAH formulation to learn hash codes for the target domain by reducing domain disparity. 
(iv) Labeled data is available in the target domain. 
This method falls under supervised hashing (\textbf{SuH}) (as it uses labeled data in the target domain to learn hash codes in the same domain) and denotes the upper bound on the performance. 
It is included to compare the performance of unsupervised hashing algorithms relative to the supervised algorithm. 
We used the DPSH algorithm \cite{li2015feature} to train a deep network on the target data and used it to generate hash codes on a validation subset. 

\begin{figure*}[t]
\centering
\subfloat[\scriptsize{Art}]{
		\label{Fig:ArtPR64}
    \includegraphics[trim = 35mm 14mm 42mm 10mm, clip, width=0.245\textwidth]{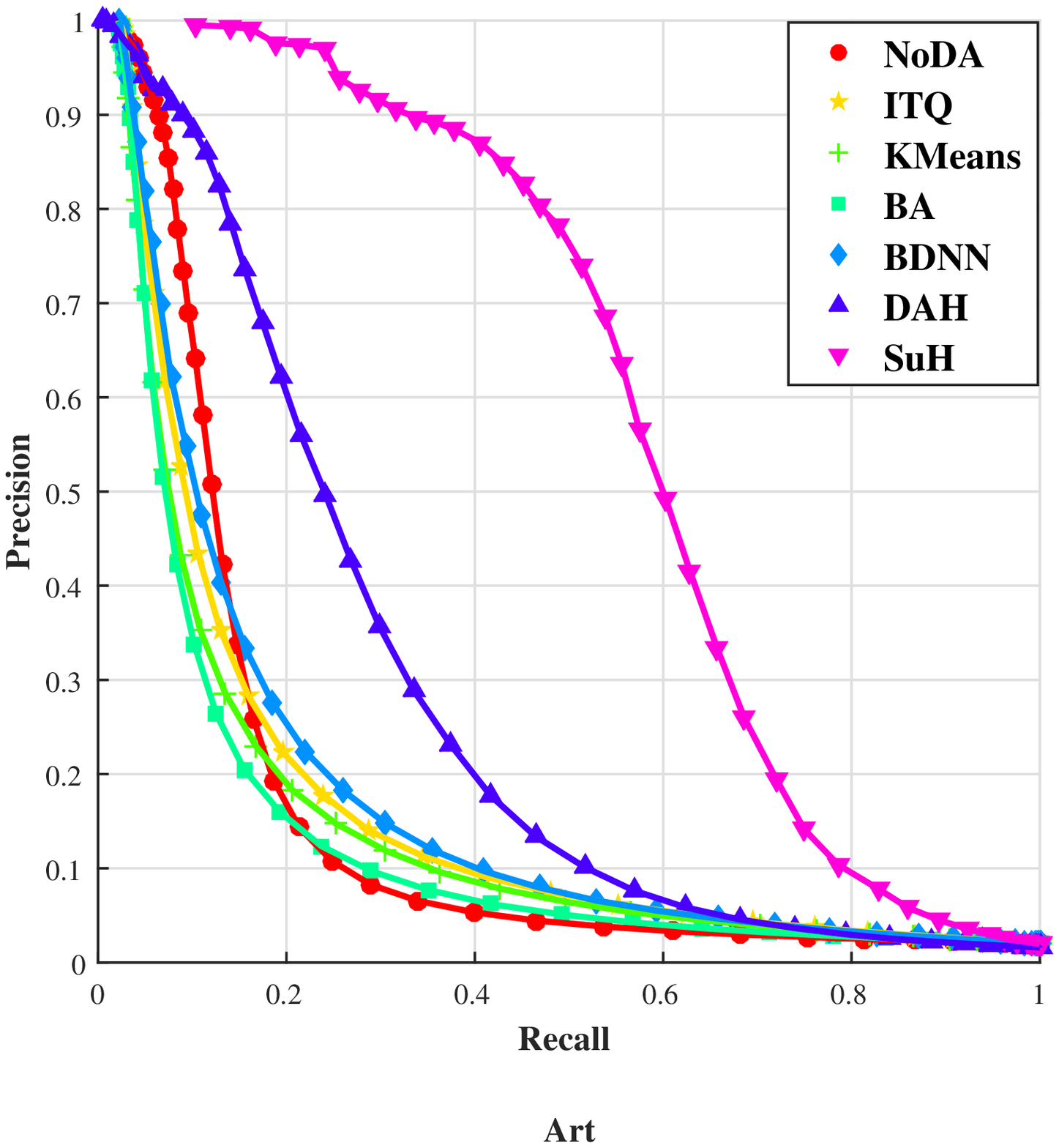}
}%
\subfloat[\scriptsize{Clipart}]{
		\label{Fig:ClipartPR64}
    \includegraphics[trim = 35mm 14mm 42mm 10mm, clip, width=0.245\textwidth]{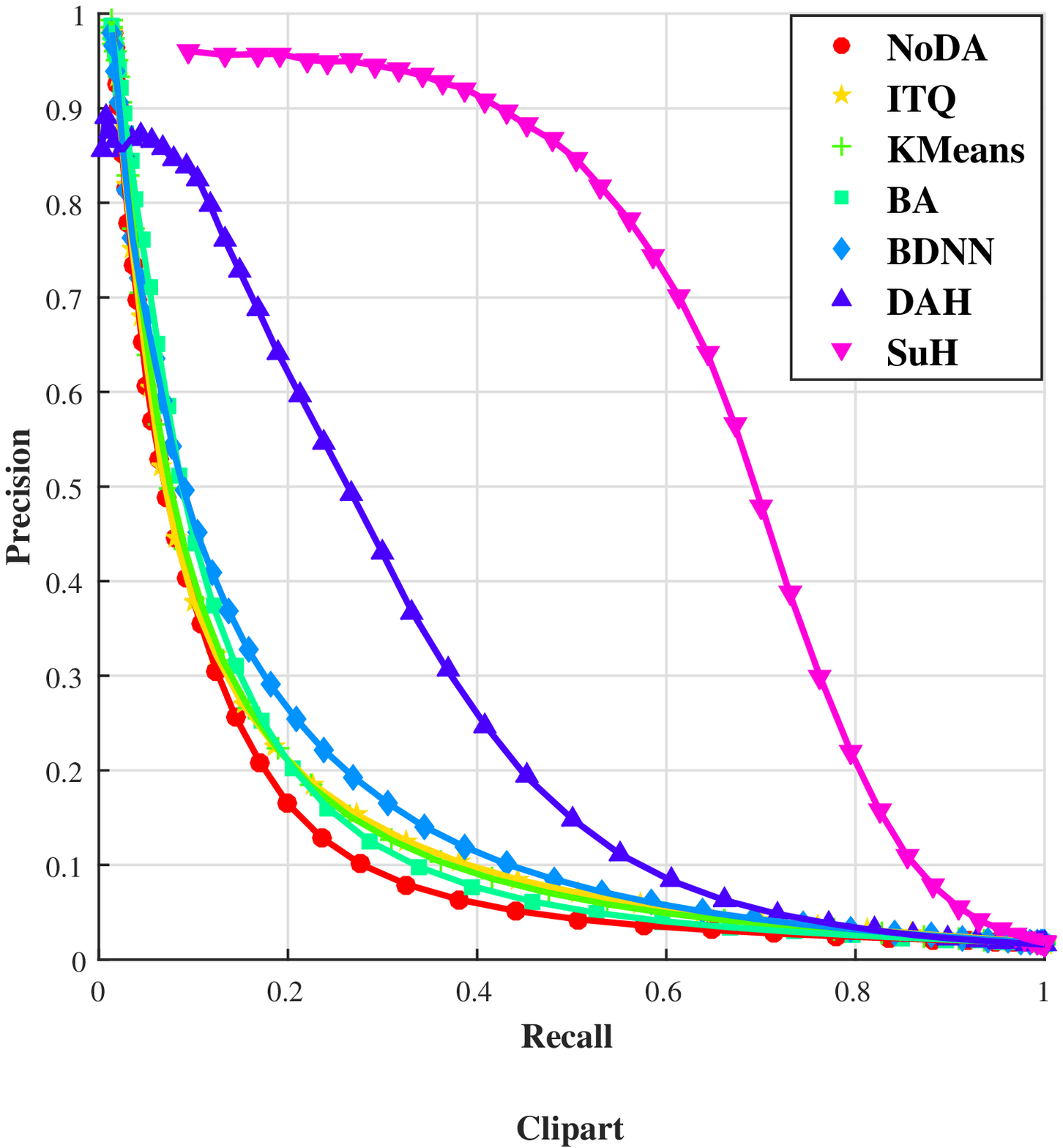}
}%
\subfloat[\scriptsize{Product}]{
		\label{Fig:ProductPR64}
    \includegraphics[trim = 35mm 14mm 42mm 10mm, clip, width=0.245\textwidth]{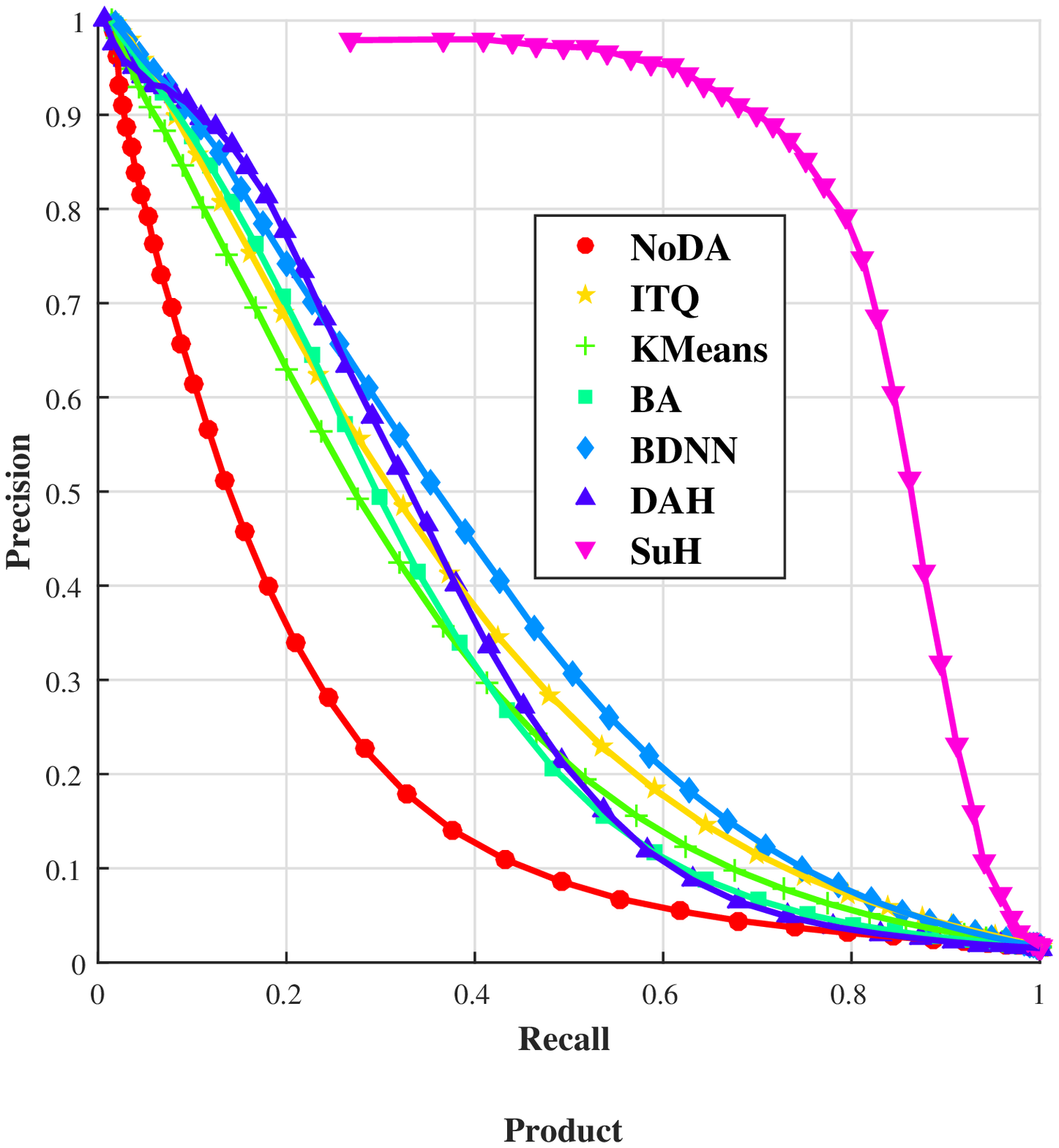}
}%
\subfloat[\scriptsize{Real-World}]{
		\label{Fig:RealWorldPR64}
    \includegraphics[trim = 35mm 14mm 42mm 10mm, clip, width=0.245\textwidth]{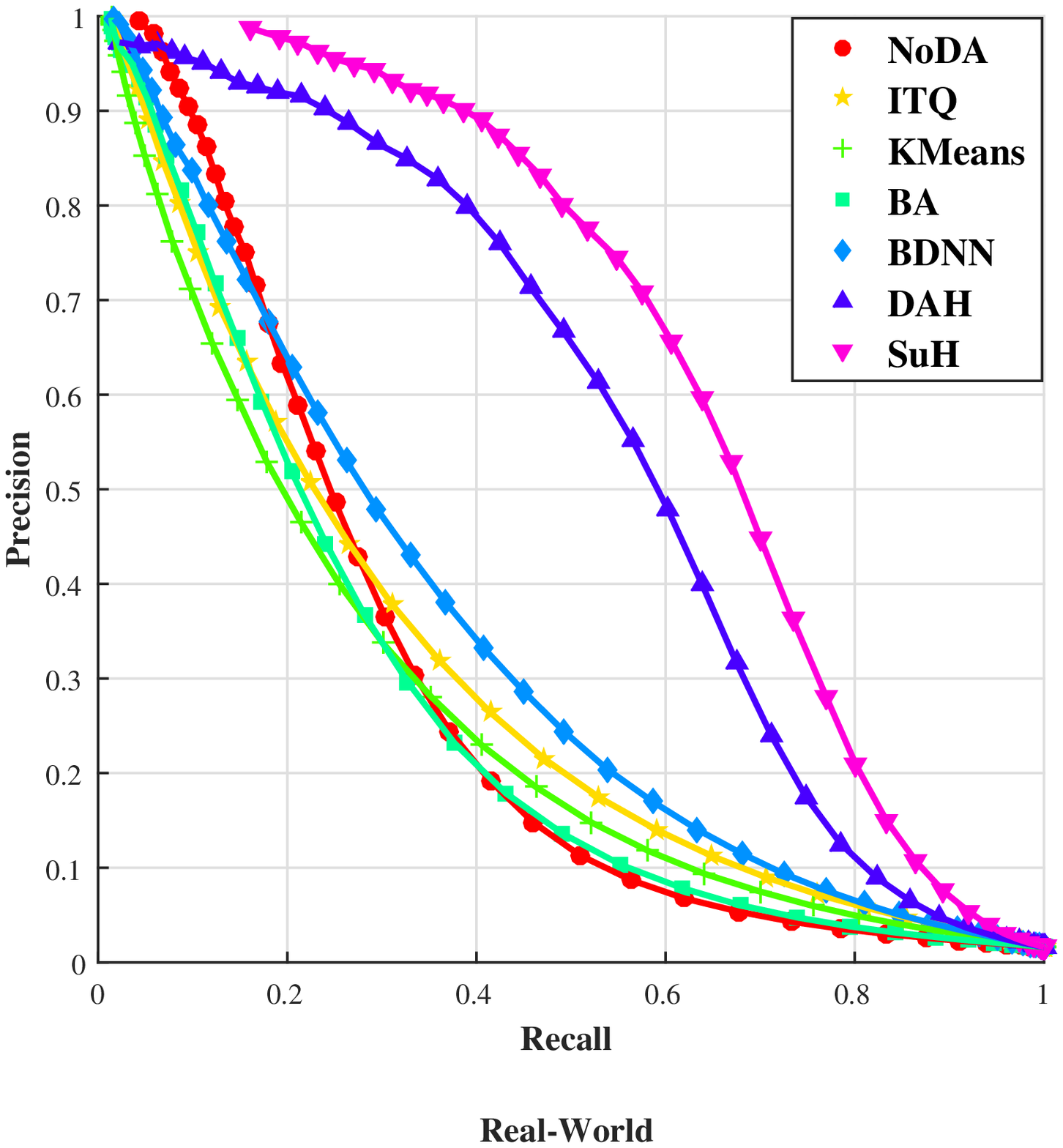}
}%
\caption{Precision-Recall curves @64 bits for the \textbf{\textit{Office-Home}} dataset. Comparison of hashing without domain adaptation (\textbf{NoDA}), shallow unsupervised hashing (\textbf{ITQ}, \textbf{KMeans}), state-of-the-art deep unsupervised hashing (\textbf{BA}, \textbf{BDNN}), unsupervised domain adaptive hashing (\textbf{DAH}) and supervised hashing (\textbf{SuH}). Best viewed in color.}
\label{Fig:OfficeHomePR64}
\end{figure*}

\begin{figure}[t]
\centering
\subfloat[\scriptsize{Amazon}]{
		\label{Fig:AmazonPR64}
    \includegraphics[trim = 35mm 14mm 42mm 10mm, clip, width=0.24\textwidth]{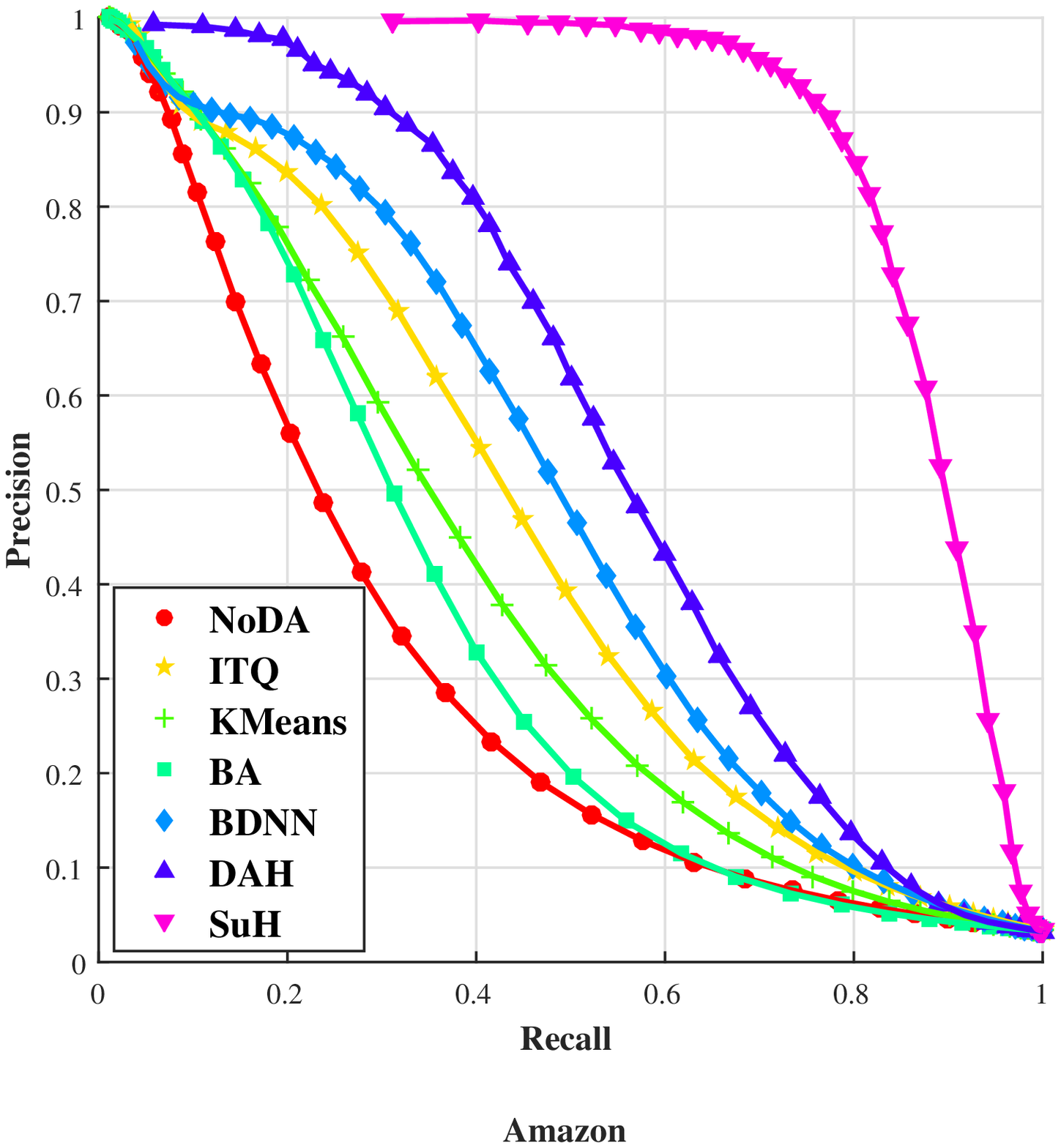}
}%
\subfloat[\scriptsize{Webcam}]{
		\label{Fig:WebcamPR64}
    \includegraphics[trim = 35mm 14mm 42mm 10mm, clip, width=0.24\textwidth]{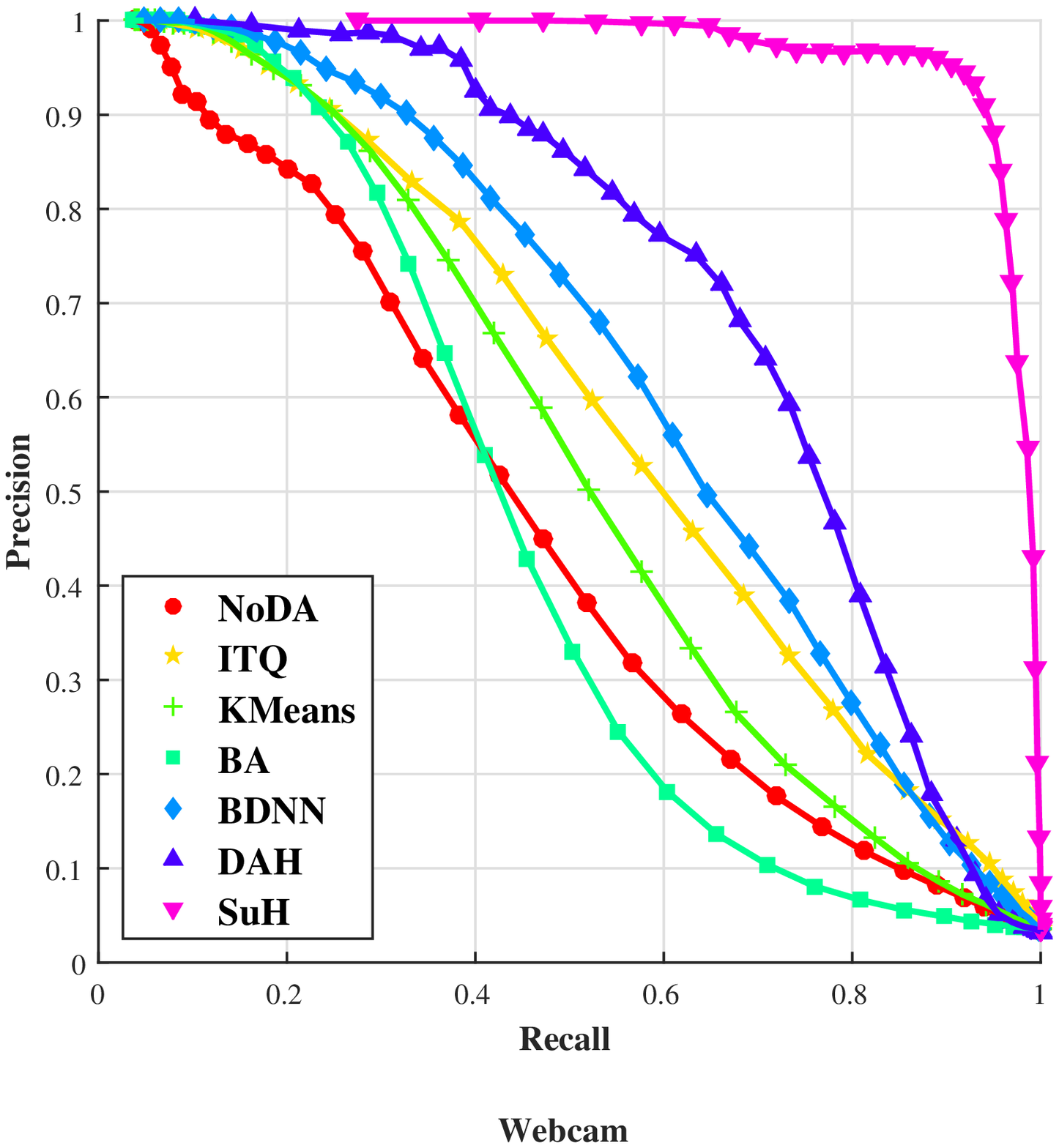}
}%
\caption{Precision-Recall curves @64 bits for the \textbf{\textit{Office}} dataset. Comparison of hashing without domain adaptation (\textbf{NoDA}), shallow unsupervised hashing (\textbf{ITQ}, \textbf{KMeans}), state-of-the-art deep unsupervised hashing (\textbf{BA}, \textbf{BDNN}), unsupervised domain adaptive hashing (\textbf{DAH}) and supervised hashing (\textbf{SuH}). Best viewed in color.}
\label{Fig:OfficePR64}
\end{figure}

\begin{table}[t]
\centering
\caption{Mean average precision @64 bits. For the NoDA and DAH results, \texttt{Art} is the source domain for \texttt{Clipart}, \texttt{Product} and \texttt{Real-World} and \texttt{Clipart} is the source domain for \texttt{Art}. Similarly, \texttt{Amazon} and \texttt{Webcam} are source target pairs.} 
\label{Tab:mAPHash64}
\resizebox{\linewidth}{!}{%
\begin{tabu}{|c|[2pt] c c c c c c c|}
\hline
 \textbf{Expt.} &  \textbf{NoDA} &  \textbf{ITQ} &  \textbf{KMeans} &  \textbf{BA} &  \textbf{BDNN} &  \textbf{DAH} & \textbf{SuH} \\\tabucline[2pt]{-}
 \texttt{Amazon}  & 0.324  & 0.465  & 0.403  & 0.367  & 0.491  & 0.582  & 0.830 \\
 \texttt{Webcam}  & 0.511  & 0.652  & 0.558  & 0.480  & 0.656  & 0.717  & 0.939 \\
 \texttt{Art}  & 0.155  & 0.191  & 0.170  & 0.156  & 0.193  & 0.302  & 0.492 \\
 \texttt{Clipart}  & 0.160  & 0.195  & 0.178  & 0.179  & 0.206  & 0.333  & 0.622 \\
 \texttt{Product}  & 0.239  & 0.393  & 0.341  & 0.349  & 0.407  & 0.414  & 0.774 \\
 \texttt{Real-World}  & 0.281  & 0.323  & 0.279  & 0.273  & 0.336  & 0.533  & 0.586 \\\hline
 Avg.  & 0.278  & 0.370  & 0.322  & 0.301  & 0.382  & 0.480  & 0.707 \\\hline
\end{tabu}
}
\end{table}

\noindent\textbf{Results and Discussion}: We applied the precision-recall curves and the mean average precision (mAP) measures to evaluate the efficacy of the hashing methods, similar to previous research \cite{carreira2015hashing, do2016learning, gong2011iterative}. 
The results are depicted in Figures \ref{Fig:OfficeHomePR64} and \ref{Fig:OfficePR64} (precision-recall curves) and Table \ref{Tab:mAPHash64} (mAP values), where we present hashing with code length $d=64$ bits. 
Hashing performance with $d=16$ bits also follows a similar trend and is presented in the supplementary material. 
For the sake of brevity, we drop the results with \texttt{Dslr} as it is very similar to \texttt{Webcam}, with little domain difference. 
We note that the NoDA has the poorest performance due to domain mismatch. 
This demonstrates that domain disparity needs to be considered before deploying a hashing network to extract hash codes. 
The unsupervised hashing methods ITQ, KMeans, BA and BDNN perform slightly better compared to NoDA. 
The proposed DAH algorithm encompasses hash code learning and domain adaptation in a single integrated framework. 
It is thus able to leverage the labeled data in the source domain in a meaningful manner to learn efficient hash codes for the target domain. 
This accounts for its improved performance, as is evident in Figures \ref{Fig:OfficeHomePR64} and \ref{Fig:OfficePR64} and Table \ref{Tab:mAPHash64}. 
The supervised hashing technique (SuH) uses labels from the target and therefore depicts the best performance. 
The proposed DAH framework consistently delivers the best performance relative to SuH when compared with the other hashing procedures. 
This demonstrates the merit of our framework in learning representative hash codes by utilizing labeled data from a different domain.  Such a framework will be immensely useful in a real-world setting. 

\section{Conclusions}
In this paper, we have proposed a novel domain adaptive hashing (DAH) framework which exploits the feature learning capabilities of deep neural networks to learn efficient hash codes for unsupervised domain adaptation. 
The DAH framework solves two important practical problems: category assignment with weak supervision or insufficient labels (through domain adaptation) and the estimation of hash codes in an unsupervised setting (hash codes for target data). 
Thus, two practical challenges are addressed through a single integrated framework.
This research is the first of its kind to integrate hash code learning with unsupervised domain adaptation. 
We also introduced a new dataset, \textit{\textbf{Office-Home}}, which can be used to further research in domain adaptation. 

\noindent\textbf{Acknowledgements}: This material is based upon work supported by the National Science Foundation (NSF) under Grant No:1116360. 
Any opinions, findings, and conclusions or recommendations expressed in this material are those of the authors and do not necessarily reflect the views of the NSF.
{\small

}

\clearpage
\onecolumn
\title{Supplementary Material}
\author{}
\date{}
\maketitle
\section{Loss Function Derivative}
\setcounter{equation}{8}
In this section we outline the derivative of Equation \ref{Eq:DAH} for the backpropagation algorithm; 
\begin{flalign}
\min_{\cU}\cJ = \cL(\cU_s) + \gamma\cM(\cU_s,\cU_t) + \eta\cH(\cU_s, \cU_t), \tag{8}
\label{Eq:DAH}
\end{flalign}
where, $\cU\coloneqq\{\cU_s\cup~\cU_t\}$ and ($\gamma$, $\eta$) control the importance of domain adaptation (\ref{Eq:MKMMD}) and target entropy loss (\ref{Eq:EntLoss}) respectively. 
In the following subsections, we outline the derivative of the individual terms w.r.t. the input $\cU$. 
\subsection{Derivative for MK-MMD}
\begin{flalign}
\cM(\cU_s,\cU_t) = \sum_{l\in\cF}d_k^2(\cU^l_s,\cU^l_t), \tag{1}
\label{Eq:MKMMD}
\end{flalign}
\begin{flalign}
d_k^2(\cU^l_s,\cU^l_t)=\Big|\Big|\bbE[\phi(\bmu^{s,l})] - \bbE[\phi(\bmu^{t,l})]\Big|\Big|_{\sH_k}^2. \tag{2}
\label{Eq:layerMKMMD}
\end{flalign}
We implement the linear MK-MMD loss according to \cite{gretton2012optimal}. 
For this derivation, we consider the loss at just one layer. 
The derivative for the MK-MMD loss at every other layer can be derived in a similar manner. 
The output of $i^{th}$ source data point at layer $l$ is represented as $\bmu_i$ and the output of the $i^{th}$ target data point is represented as $\bmv_i$. 
For ease of representation, we drop the superscripts for the source $(s)$, the target $(t)$ and the layer $(l)$. 
Unlike the conventional MMD loss which is $\cO(n^2)$, the MK-MMD loss outlined in \cite{gretton2012optimal} is $\cO(n)$ and can be estimated online (does not require all the data). 
The loss is calculated over every batch of data points during the back-propagation. 
Let $n$ be the number of source data points $\cU \coloneqq \{\bmu_i\}_{i=1}^n$ and the number of target data points $\cV \coloneqq \{\bmv_i\}_{i=1}^n$ in the batch. 
We assume equal number of source and target data points in a batch and that $n$ is even. 
The MK-MMD is defined over a set of $4$ data points $\bmw_i = [\bmu_{2i-1},\bmu_{2i},\bmv_{2i-1},\bmv_{2i}], \forall i\in\{1,2,\ldots,n/2\}$. 
The MK-MMD is given by,
\begin{flalign}
\cM(\cU,\cV) = \sum_{m=1}^\kappa\beta_m\frac{1}{n/2}\sum_{i=1}^{n/2}h_m(\bmw_i),
\label{Eq:MKMMDDer1}
\end{flalign}
where, $\kappa$ is the number of kernels and $\beta_m = 1/\kappa$ is the weight for each kernel and,
\begin{flalign}
h_m(\bmw_i) = k_m(\bmu_{2i-1},\bmu_{2i}) + k_m(\bmv_{2i-1},\bmv_{2i}) - k_m(\bmu_{2i-1},\bmv_{2i}) - k_m(\bmu_{2i},\bmv_{2i-1}),
\label{Eq:MKMMDDer-hm}
\end{flalign}
where, $k_m(\bmx, \bmy) = \text{exp}\big(-\frac{||\bmx -\bmy||_2^2}{\sigma_m}\big)$. 
Re-writing the MK-MMD in terms of the kernels, we have,
\begin{flalign}
\cM(\cU,\cV) = \frac{2}{n\kappa}\sum_{m=1}^\kappa\sum_{i=1}^{n/2}\big[k_m(\bmu_{2i-1},\bmu_{2i}) + k_m(\bmv_{2i-1},\bmv_{2i}) - k_m(\bmu_{2i-1},\bmv_{2i}) - k_m(\bmu_{2i},\bmv_{2i-1})\big],
\label{Eq:MKMMDDer-km}
\end{flalign}
We now outline the derivative of \ref{Eq:MKMMDDer-km} w.r.t. source output $\bmu_q$ and target output $\bmv_q$. The derivative is,
\begin{flalign}
\frac{\partial\cM}{\partial\bmu_q} = \frac{2}{n\kappa}\sum_{m=1}^\kappa\sum_{i=1}^{n/2}\Big[
&\frac{2}{\sigma_m}k_m(\bmu_{2i-1},\bmu_{2i}).(\bmu_{2i-1}-\bmu_{2i}).(\cI\{q=2i\} - \cI\{q=2i-1\}) \notag \\
&+ \frac{2}{\sigma_m}k_m(\bmu_{2i-1},\bmv_{2i}).(\bmu_{2i-1}-\bmv_{2i}).\cI\{q=2i-1\} 
+ \frac{2}{\sigma_m}k_m(\bmu_{2i},\bmv_{2i-1}).(\bmu_{2i}-\bmv_{2i-1}).\cI\{q=2i\}
\Big],
\label{Eq:MKMMDDer-uq}
\end{flalign}
where, $\cI\{.\}$ is the indicator function which is $1$ if the condition is true, else it is false. 
The derivative w.r.t. the target data output $\bmv_q$ is,
\begin{flalign}
\frac{\partial\cM}{\partial\bmv_q} = \frac{2}{n\kappa}\sum_{m=1}^\kappa\sum_{i=1}^{n/2}\Big[
&\frac{2}{\sigma_m}k_m(\bmv_{2i-1},\bmv_{2i}).(\bmv_{2i-1}-\bmv_{2i}).(\cI\{q=2i\} - \cI\{q=2i-1\}) \notag \\
&- \frac{2}{\sigma_m}k_m(\bmu_{2i-1},\bmv_{2i}).(\bmu_{2i-1}-\bmv_{2i}).\cI\{q=2i\} 
- \frac{2}{\sigma_m}k_m(\bmu_{2i},\bmv_{2i-1}).(\bmu_{2i}-\bmv_{2i-1}).\cI\{q=2i-1\}
\Big],
\label{Eq:MKMMDDer-vq}
\end{flalign}

\subsection{Derivative for Supervised Hash Loss}
The supervised hash loss is given by,
\begin{flalign}
\min_{\cU_s} \cL(\cU_s) =& -\sum_{s_{ij} \in \cS}\Big( s_{ij}\bmu_i^\top \bmu_j - \text{log}\big(1 + \text{exp}(\bmu_i^\top \bmu_j)\big)\Big)\notag\\
		& + \sum_{i=1}^{n_s}\big|\big|\bmu_i - \text{sgn}(\bmu_i)\big|\big|_2^2. \tag{5}
\label{Eq:supHashLoss}
\end{flalign}
The partial derivative of \ref{Eq:supHashLoss} w.r.t. source data output $\bmu_p$ is given by,
\begin{flalign}
\frac{\partial\cL}{\partial\bmu_q} = \sum_{s_{ij}\in\cS}\Big[I\{i=q\}\big(\sigma(\bmu_i^\top\bmu_j) -s_{ij}\big)\bmu_j +  I\{j=q\}\big(\sigma(\bmu_i^\top\bmu_j) -s_{ij}\big)\bmu_i\Big] + 2(\bmu_q - \text{sgn}(\bmu_q)) \,
\label{Eq:supHashLossDer1}
\end{flalign}
where, $\sigma(x) = \frac{1}{1+exp(-x)}$. We assume $\text{sgn}(.)$ to be a constant and avoid the differentiability issues with $\text{sgn}(.)$ at $0$. 
Since the $\cS$ is symmetric, we can reduce the derivative to,
\begin{flalign}
\frac{\partial\cL}{\partial\bmu_q} = \sum_{j=1}^{n_s}\Big[2\big(\sigma(\bmu_q^\top\bmu_j) -s_{qj}\big)\bmu_j\Big] + 2\big(\bmu_q - \text{sgn}(\bmu_q)\big) .
\label{Eq:supHashLossDer}
\end{flalign}
\subsection{Derivative for Unsupervised Entropy Loss}
We outline the derivative of $\frac{d\cH}{d\cU}$ in the following section, where $\cH$ is defined as,
\begin{flalign}
\cH(\cU_s, \cU_t) = -\frac{1}{n_t}\sum_{i = 1}^{n_t}\sum_{j=1}^C p_{ij}\text{log}(p_{ij}) \tag{7}
\label{Eq:EntLoss}
\end{flalign}
and $p_{ij}$ is the probability of target data output $\bmu_i^t$ belonging to category $j$, given by
\begin{flalign}
p_{ij} = \frac{\sum_{k=1}^K \text{exp}({\bmu_i^t}^\top \bmu_k^{s_j})}{\sum_{l=1}^C\sum_{k'=1}^K \text{exp}({\bmu_i^t}^\top \bmu_{k'}^{s_l})} \tag{6}
\label{Eq:EntProb}
\end{flalign}
For ease of representation, we will denote the target output $\bmu_i^t$ as $\bmv_i$ and drop the superscript $t$. 
Similarly, we will denote the $k^{th}$ source data point in the $j^{th}$ category $\bmu_k^{s_j}$ as $\bmu_k^j$, by dropping the domain superscript. We define the probability $p_{ij}$ with the news terms as,
\begin{flalign}
p_{ij} = \frac{\sum_{k=1}^K \text{exp}({\bmv_i}^\top \bmu_k^j)}{\sum_{l=1}^C\sum_{k'=1}^K \text{exp}({\bmv_i}^\top \bmu_{k'}^l)}
\label{Eq:EntNewProb}
\end{flalign}
Further, we simplify by replacing $\text{exp}(\bmv_i^\top\bmu_k^j)$ with $\text{exp}(i,jk)$. Equation \ref{Eq:EntNewProb} can now be represented as,
\begin{flalign}
p_{ij} = \frac{\sum_{k=1}^K \text{exp}(i, jk)}{\sum_{l=1}^C\sum_{k'=1}^K \text{exp}(i, lk')}
\label{Eq:EntShortHandProb}
\end{flalign}
We drop the outer summations (along with the -ve sign) and will reintroduce it at a later time. The entropy loss can be re-phrased using log($\frac{a}{b}$) = log($a$) - log($b$) as, 
\begin{flalign}
\cH_{ij}= &\frac{\sum_{k=1}^K\text{exp}(i, jk)}{\sum_{l=1}^C\sum_{k'=1}^K \text{exp}(i, lk')} \text{log}\big( \textstyle \sum_{k=1}^K\text{exp}(i, jk)\big) \, \label{Eq:EntShortHandProb1}\\
&-\frac{\sum_{k=1}^K\text{exp}(i, jk)}{\sum_{l=1}^C\sum_{k'=1}^K \text{exp}(i, lk')} \text{log}\big(\textstyle\sum_{l=1}^C\textstyle\sum_{k'=1}^K \text{exp}(i, lk')\big)
\label{Eq:EntShortHandProb2}
\end{flalign}
We need to estimate both, $\frac{\partial\cH_{ij}}{\partial\bmv_i}$ for the target and $\frac{\partial\cH_{ij}}{\partial\bmu_q^p}$ for the source. 
We refer to $\partial\bmu_q^p$ for a consistent reference to source data. 
The derivative $\frac{\partial\cH_{ij}}{\partial\bmu_q^p}$ for \ref{Eq:EntShortHandProb1} is,
\begin{flalign}
\bigg[\frac{\partial\cH_{ij}}{\partial\bmu_q^p}\bigg]_{\ref{Eq:EntShortHandProb1}} = \frac{\bmv_i}{\sum_{l,k'}\text{exp}(i,lk')}\Big[&\textstyle\sum_k I\{\substack{j=p,\\k=q}\}\text{exp}(i,jk).\text{log}\big(\textstyle\sum_k\text{exp}(i,jk)\big) + \textstyle\sum_k I\{\substack{j=p,\\k=q}\}\text{exp}(i,jk)  \notag\\
&- p_{ij}\text{exp}(i,pq) \text{log}\big(\textstyle\sum_k\text{exp}(i,jk)\big) \Big],
\label{Eq:EntDerU1}
\end{flalign}
where, $I\{.\}$ is an indicator function which is $1$ only when both the conditions within are true, else it is $0$. 
The derivative $\frac{\partial\cH_{ij}}{\partial\bmu_q^p}$ for \ref{Eq:EntShortHandProb2} is,
\begin{flalign}
\bigg[\frac{\partial\cH_{ij}}{\partial\bmu_q^p}\bigg]_{\ref{Eq:EntShortHandProb2}} = -\frac{\bmv_i}{\sum_{l,k'}\text{exp}(i,lk')}\Big[& \textstyle\sum_k I\{\substack{j=p,\\k=q}\}\text{exp}(i,jk).\text{log}\big(\textstyle\sum_{l,k'}\text{exp}(i,lk')\big) + p_{ij}\text{exp}(i,pq) \notag\\
&- p_{ij}\text{exp}(i,pq) \text{log}\big(\textstyle\sum_{l,k'}\text{exp}(i,lk')\big) \Big]
\label{Eq:EntDerU2}
\end{flalign}
Expressing $\frac{\partial\cH_{ij}}{\partial\bmu_q^p} = \Big[\frac{\partial\cH_{ij}}{\partial\bmu_q^p}\Big]_{\ref{Eq:EntShortHandProb1}}+\Big[\frac{\partial\cH_{ij}}{\partial\bmu_q^p}\Big]_{\ref{Eq:EntShortHandProb2}}$, and defining $\bar{p}_{ijk} = \frac{\text{exp}(i,jk)}{\sum_{l,k'}\text{exp}(i,lk')}$ the derivative w.r.t. the source is, 
\begin{flalign}
\frac{\partial\cH_{ij}}{\partial\bmu_q^p} 
=&\bmv_i\Big[ 
\textstyle\sum_k I\{\substack{j=p,\\k=q}\}\bar{p}_{ijk}.\text{log}\big(\textstyle\sum_k\text{exp}(i,jk)\big) + \textstyle\sum_k I\{\substack{j=p,\\k=q}\}\bar{p}_{ijk}  \notag\\
&- p_{ij}\bar{p}_{ipq} \text{log}\big(\textstyle\sum_k\text{exp}(i,jk)\big) -\textstyle\sum_k I\{\substack{j=p,\\k=q}\}\bar{p}_{ijk}.\text{log}\big(\textstyle\sum_{l,k'}\text{exp}(i,lk')\big) \notag\\
&- p_{ij}\bar{p}_{ipq} + p_{ij}\bar{p}_{ipq} \text{log}\big(\textstyle\sum_{l,k'}\text{exp}(i,lk')\big)
\Big]\\
=&\bmv_i \Big[\textstyle\sum_k I\{\substack{j=p,\\k=q}\}\bar{p}_{ijk}\text{log}(p_{ij}) - p_{ij}\bar{p}_{ipq}\text{log}(p_{ij}) + \textstyle\sum_k I\{\substack{j=p,\\k=q}\}\bar{p}_{ijk} -p_{ij}\bar{p}_{ipq} \Big]\\
=&\bmv_i\big(\text{log}(p_{ij}) + 1\big)\Big[\textstyle\sum_k I\{\substack{j=p,\\k=q}\}\bar{p}_{ijk} - p_{ij}\bar{p}_{ipq} \Big]
\label{Eq:EntDerUij}
\end{flalign}
The derivative of $\cH$ w.r.t the \textbf{source} output $\bmu_q^p$ is given by,
\begin{flalign}
\frac{\partial\cH}{\partial\bmu_q^p} = -\frac{1}{n_t}\sum_{i=1}^{n_t}\sum_{j=1}^C\bmv_i\big(\text{log}(p_{ij}) + 1\big)\Big[\textstyle\sum_k I\{\substack{j=p,\\k=q}\}\bar{p}_{ijk} - p_{ij}\bar{p}_{ipq} \Big]
\label{Eq:EntDerU}
\end{flalign}
We now outline the derivative  $\frac{\partial\cH}{\partial\bmv_i}$ for \ref{Eq:EntShortHandProb1} as,
\begin{flalign}
\bigg[\frac{\partial\cH_{ij}}{\partial\bmv_i}\bigg]_{\ref{Eq:EntShortHandProb1}} = \frac{1}{\sum_{l,k'}\text{exp}(i,lk')}\Big[
& \text{log}\big(\textstyle\sum_k\text{exp}(i,jk)\big)\sum_k\text{exp}(i,jk)\bmu_k^j + \sum_k\text{exp}(i,jk)\bmu_k^j \notag\\
&- \frac{1}{\sum_{l,k'}\text{exp}(i,lk')}\textstyle\sum_k\text{exp}(i,jk)\text{log}\big(\textstyle\sum_k\text{exp}(i,jk)\big)\textstyle\sum_{l,k'}\text{exp}(i,lk')\bmu_{k'}^l 
\Big],
\label{Eq:EntDerV1}
\end{flalign}
and the derivative $\frac{\partial\cH}{\partial\bmv_i}$ for \ref{Eq:EntShortHandProb2} as,
\begin{flalign}
\bigg[\frac{\partial\cH_{ij}}{\partial\bmv_i}\bigg]_{\ref{Eq:EntShortHandProb2}} = -\frac{1}{\sum_{l,k'}\text{exp}(i,lk')}\Big[
& \text{log}\big(\textstyle\sum_{l,k'}\text{exp}(i,lk')\big)\sum_k\text{exp}(i,jk)\bmu_k^j + \frac{\sum_k\text{exp}(i,jk)}{\sum_{l,k'}\text{exp}(i,lk')}\sum_{l,k'}\text{exp}(i,lk')\bmu_{k'}^l \notag\\
&- \frac{1}{\sum_{l,k'}\text{exp}(i,lk')}\textstyle\sum_k\text{exp}(i,jk)\text{log}\big(\textstyle\sum_{l,k'}\text{exp}(i,lk')\big)\textstyle\sum_{l,k'}\text{exp}(i,lk')\bmu_{k'}^l 
\Big],
\label{Eq:EntDerV2}
\end{flalign}
Expressing $\frac{\partial\cH_{ij}}{\partial\bmv_i} = \Big[\frac{\partial\cH_{ij}}{\partial\bmv_i}\Big]_{\ref{Eq:EntShortHandProb1}}+\Big[\frac{\partial\cH_{ij}}{\partial\bmv_i}\Big]_{\ref{Eq:EntShortHandProb2}}$, we get,
\begin{flalign}
\frac{\partial\cH_{ij}}{\partial\bmv_i} &= \frac{1}{\sum_{l,k'}\text{exp}(i,lk')}\Big[
\text{log}\big(\textstyle\sum_k\text{exp}(i,jk)\big)\sum_k\text{exp}(i,jk)\bmu_k^j - \text{log}\big(\textstyle\sum_{l,k'}\text{exp}(i,lk')\big)\sum_k\text{exp}(i,jk)\bmu_k^j \notag \\
& \quad + \textstyle\sum_k\text{exp}(i,jk)\bmu_k^j -  p_{ij}\sum_{l,k'}\text{exp}(i,lk')\bmu_{k'}^l \notag \\
& \quad - p_{ij}\text{log}\big(\textstyle\sum_k\text{exp}(i,jk)\big)\textstyle\sum_{l,k'}\text{exp}(i,lk')\bmu_{k'}^l 
+ p_{ij}\text{log}\big(\textstyle\sum_{l,k'}\text{exp}(i,lk')\big)\textstyle\sum_{l,k'}\text{exp}(i,lk')\bmu_{k'}^l
\Big]\\
&=\Big[
\text{log}\big(\textstyle\sum_k\text{exp}(i,jk)\big)\sum_k\bar{p}_{ijk}\bmu_k^j - \text{log}\big(\textstyle\sum_{l,k'}\text{exp}(i,lk')\big)\sum_k\bar{p}_{ijk}\bmu_k^j \notag \\
& \quad + \textstyle\sum_k\bar{p}_{ijk}\bmu_k^j -  p_{ij}\sum_{l,k'}\bar{p}_{ijk'}\bmu_{k'}^l \notag \\
& \quad - p_{ij}\text{log}\big(\textstyle\sum_k\text{exp}(i,jk)\big)\textstyle\sum_{l,k'}\bar{p}_{ijk'}\bmu_{k'}^l 
+ p_{ij}\text{log}\big(\textstyle\sum_{l,k'}\text{exp}(i,lk')\big)\textstyle\sum_{l,k'}\bar{p}_{ijk'}\bmu_{k'}^l
\Big]\\
&=\big(\text{log}(p_{ij}) + 1\big)\textstyle\sum_k\bar{p}_{ijk}\bmu_k^j - \big(\text{log}(p_{ij}) + 1\big)p_{ij}\sum_{l,k'}\bar{p}_{ijk'}\bmu_{k'}^l\\
&=\big(\text{log}(p_{ij}) + 1\big)\big( \textstyle\sum_k\bar{p}_{ijk}\bmu_k^j - p_{ij}\sum_{l,k'}\bar{p}_{ijk'}\bmu_{k'}^l\big)
\label{Eq:EntDerVi}
\end{flalign}
The derivative of $\cH$ w.r.t. \textbf{target} output $\bmv_q$ is given by,
\begin{flalign}
\frac{\partial\cH}{\partial\bmv_q} = -\frac{1}{n_t}\sum_{j=1}^C\big(\text{log}(p_{qj}) + 1\big)\big( \textstyle\sum_k\bar{p}_{qjk}\bmu_k^j - p_{qj}\sum_{l,k'}\bar{p}_{qjk'}\bmu_{k'}^l\big)
\label{Eq:EntDerV}
\end{flalign}
The derivative of $\cH$ w.r.t. the source outputs is given by \ref{Eq:EntDerU} and w.r.t. the target outputs is given by \ref{Eq:EntDerV}. 

\section{Unsupervised Domain Adaptation: Additional Results}
In the main paper we had presented results for unsupervised domain adaptation based object recognition with $d=64$ bits. 
Here, we outline the classification results with $d=16$ (DAH-16) and $d=128$ (DAH-128) bits for the \textit{Office-Home} dataset in Table \ref{Tab:OfficeHomeAccDAH16128}. 
We also present the (DAH-64), DAN and DANN results for comparison. 
There is an increase in the average recognition accuracy for $d=128$ bits compared to $d=64$ bits because of the increased capacity in representation. 
As expected, $d=16$ has a lower recognition accuracy. 

\begin{table*}[!ht]
\centering
\caption{Recognition accuracies (\%) for domain adaptation experiments on the \textit{Office-Home} dataset. \{\texttt{Art} (Ar), \texttt{Clipart} (Cl), \texttt{Product} (Pr), \texttt{Real-World} (Rw)\}. Ar$\rightarrow$Cl implies Ar is source and Cl is target.}
\label{Tab:OfficeHomeAccDAH16128}
\resizebox{\textwidth}{!}{%
\begin{tabu}{c |[2pt] c c c c c c c c c c c c | c}
\hline
\textbf{Expt.} &  \textbf{Ar$\rightarrow$Cl} &  \textbf{Ar$\rightarrow$Pr} &  \textbf{Ar$\rightarrow$Rw} &  \textbf{Cl$\rightarrow$Ar} &  \textbf{Cl$\rightarrow$Pr} &  \textbf{Cl$\rightarrow$Rw} &  \textbf{Pr$\rightarrow$Ar} &  \textbf{Pr$\rightarrow$Cl} &  \textbf{Pr$\rightarrow$Rw} &  \textbf{Rw$\rightarrow$Ar} &  \textbf{Rw$\rightarrow$Cl} &  \textbf{Rw$\rightarrow$Pr} & \textbf{Avg.} \\\tabucline[2pt]{-}
 DAN  & 30.66  & 42.17  & 54.13  & 32.83  & 47.59  & 49.78  & 29.07  & 34.05  & 56.70  & 43.58  & 38.25  & 62.73  & 43.46 \\
 DANN  & 33.33  & 42.96  & 54.42  & 32.26  & 49.13  & 49.76  & 30.49  & 38.14  & 56.76  & 44.71  & 42.66  & 64.65  & 44.94 \\\hline
 DAH-16	& 23.83	& 30.32	& 40.14	& 25.67	& 38.79	& 33.26	& 20.11	& 27.72	& 40.90	& 32.63	& 25.54	& 37.46	& 31.36\\
 DAH-64	& 31.64	& 40.75	& 51.73	& 34.69	& 51.93	& 52.79	& 29.91	& 39.63	& 60.71	& 44.99	& 45.13	& 62.54	& 45.54\\
 DAH-128 & 32.58 & 40.64 & 52.40 & 35.72 & 52.80 & 52.12 & 30.94 & 41.31 & 59.31 & 45.65 & 46.67 & 64.97 & 46.26\\\hline
\end{tabu}
}
\end{table*}

\section{Unsupervised Domain Adaptive Hashing: Additional Results}

\begin{table}[t]
\centering
\caption{Mean average precision @16 bits. For the NoDA and DAH results, \texttt{Art} is the source domain for \texttt{Clipart}, \texttt{Product} and \texttt{Real-World} and \texttt{Clipart} is the source domain for \texttt{Art}. } 
\label{Tab:mAPHash16}
\resizebox{0.6\linewidth}{!}{%
\begin{tabu}{|c|[2pt] c c c c c c c|}
\hline
 \textbf{Expt.} &  \textbf{NoDA} &  \textbf{ITQ} &  \textbf{KMeans} &  \textbf{BA} &  \textbf{BDNN} &  \textbf{DAH} & \textbf{SuH} \\\tabucline[2pt]{-}
 \texttt{Art} & 0.102 & 0.147 & 0.133 & 0.131 & 0.151 & 0.207 & 0.381 \\
 \texttt{Clipart} & 0.110 & 0.120 & 0.116 & 0.123 & 0.138 & 0.211 & 0.412 \\
 \texttt{Product} & 0.134 & 0.253 & 0.241 & 0.253 & 0.313 & 0.257 & 0.459 \\
 \texttt{Real-World} & 0.193 & 0.225 & 0.195 & 0.216	& 0.248 & 0.371 & 0.400 \\\hline
 Avg. & 0.135 & 0.186 & 0.171 & 0.181 & 0.212 & 0.262 & 0.413 \\\hline
\end{tabu}
}
\end{table}

\begin{table}[t]
\centering
\caption{Mean average precision @128 bits. For the NoDA and DAH results, \texttt{Art} is the source domain for \texttt{Clipart}, \texttt{Product} and \texttt{Real-World} and \texttt{Clipart} is the source domain for \texttt{Art}. } 
\label{Tab:mAPHash128}
\resizebox{0.6\linewidth}{!}{%
\begin{tabu}{|c|[2pt] c c c c c c c|}
\hline
 \textbf{Expt.} &  \textbf{NoDA} &  \textbf{ITQ} &  \textbf{KMeans} &  \textbf{BA} &  \textbf{BDNN} &  \textbf{DAH} & \textbf{SuH} \\\tabucline[2pt]{-}
 \texttt{Art}        & 0.154 & 0.202 & 0.175 & 0.148 & 0.207 & 0.314 & 0.444 \\
 \texttt{Clipart}    & 0.186 & 0.210 & 0.196 & 0.187 & 0.213 & 0.350 & 0.346 \\
 \texttt{Product}    & 0.279 & 0.416 & 0.356 & 0.336 & 0.432 & 0.424 & 0.792 \\
 \texttt{Real-World} & 0.308 & 0.343 & 0.289 & 0.258 & 0.348 & 0.544 & 0.458 \\\hline
 Avg.                & 0.232 & 0.293 & 0.254 & 0.232 & 0.300 & 0.408 & 0.510 \\\hline
\end{tabu}
}
\end{table}

\begin{figure*}[!ht]
\centering
\subfloat[\scriptsize{Art}]{
		\label{Fig:ArtPR16}
    \includegraphics[trim = 35mm 14mm 38mm 10mm, clip, width=0.245\textwidth]{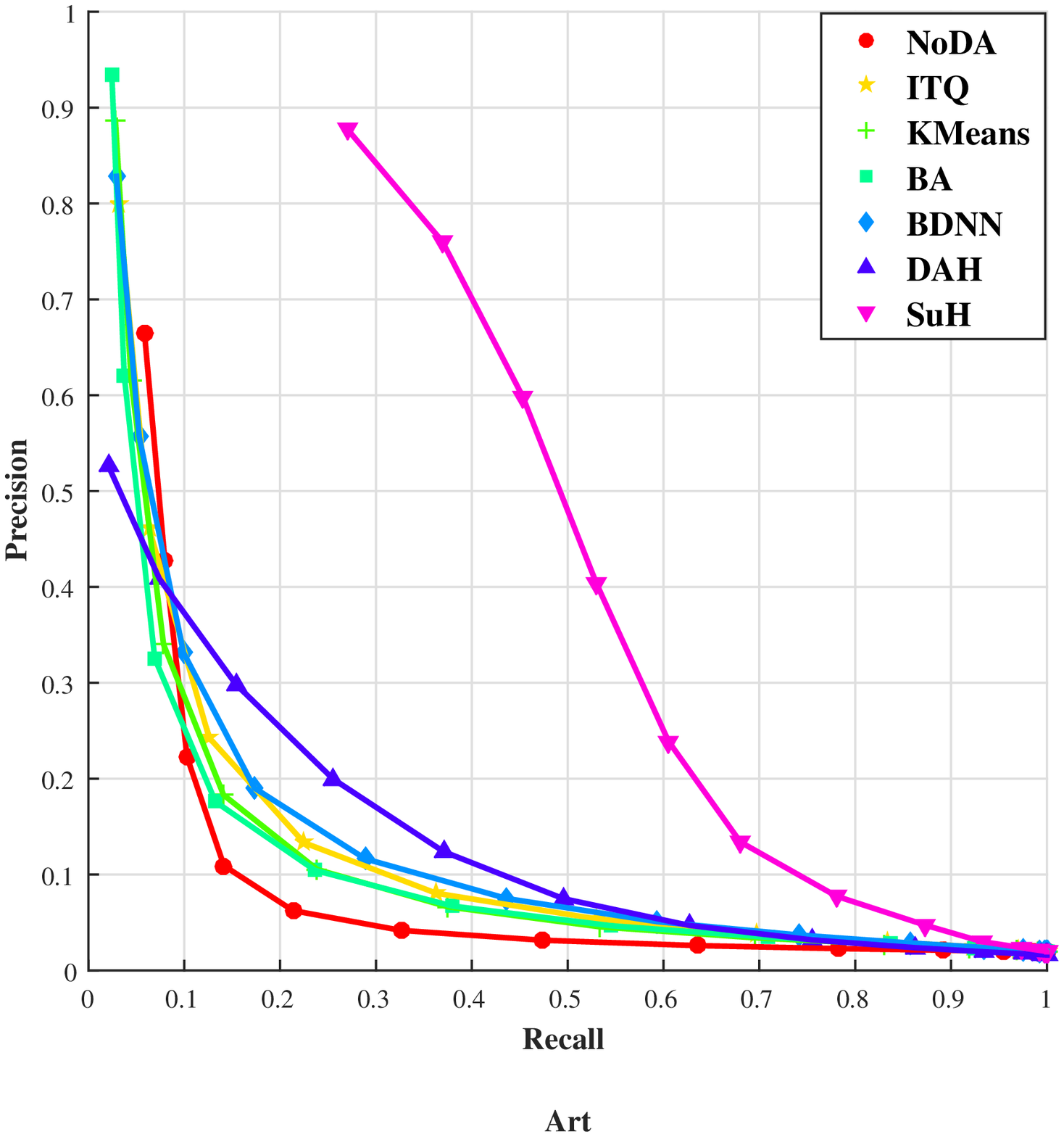}
}%
\subfloat[\scriptsize{Clipart}]{
		\label{Fig:ClipartPR16}
    \includegraphics[trim = 35mm 14mm 38mm 10mm, clip, width=0.245\textwidth]{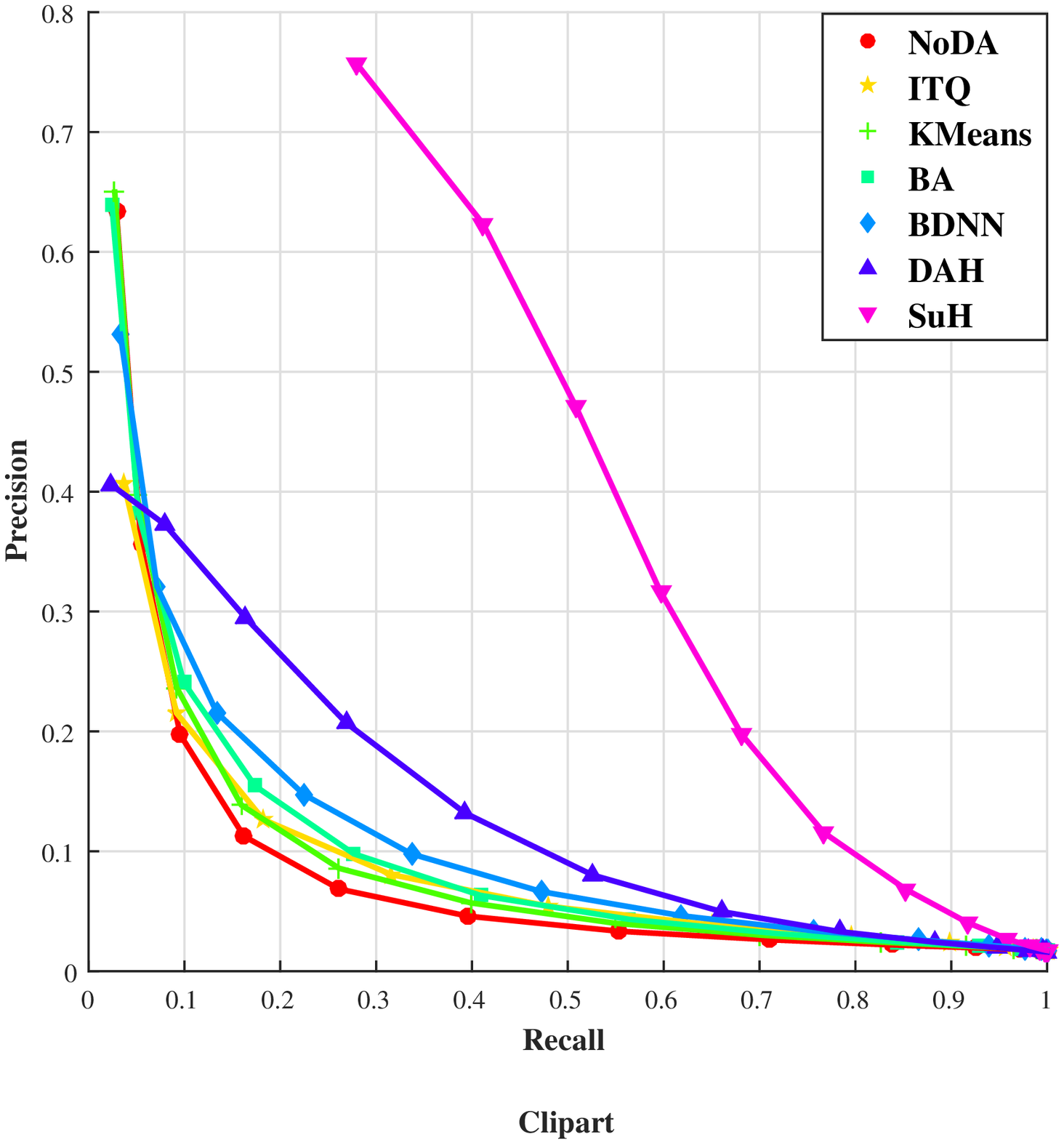}
}%
\subfloat[\scriptsize{Product}]{
		\label{Fig:ProductPR16}
    \includegraphics[trim = 35mm 14mm 38mm 10mm, clip, width=0.245\textwidth]{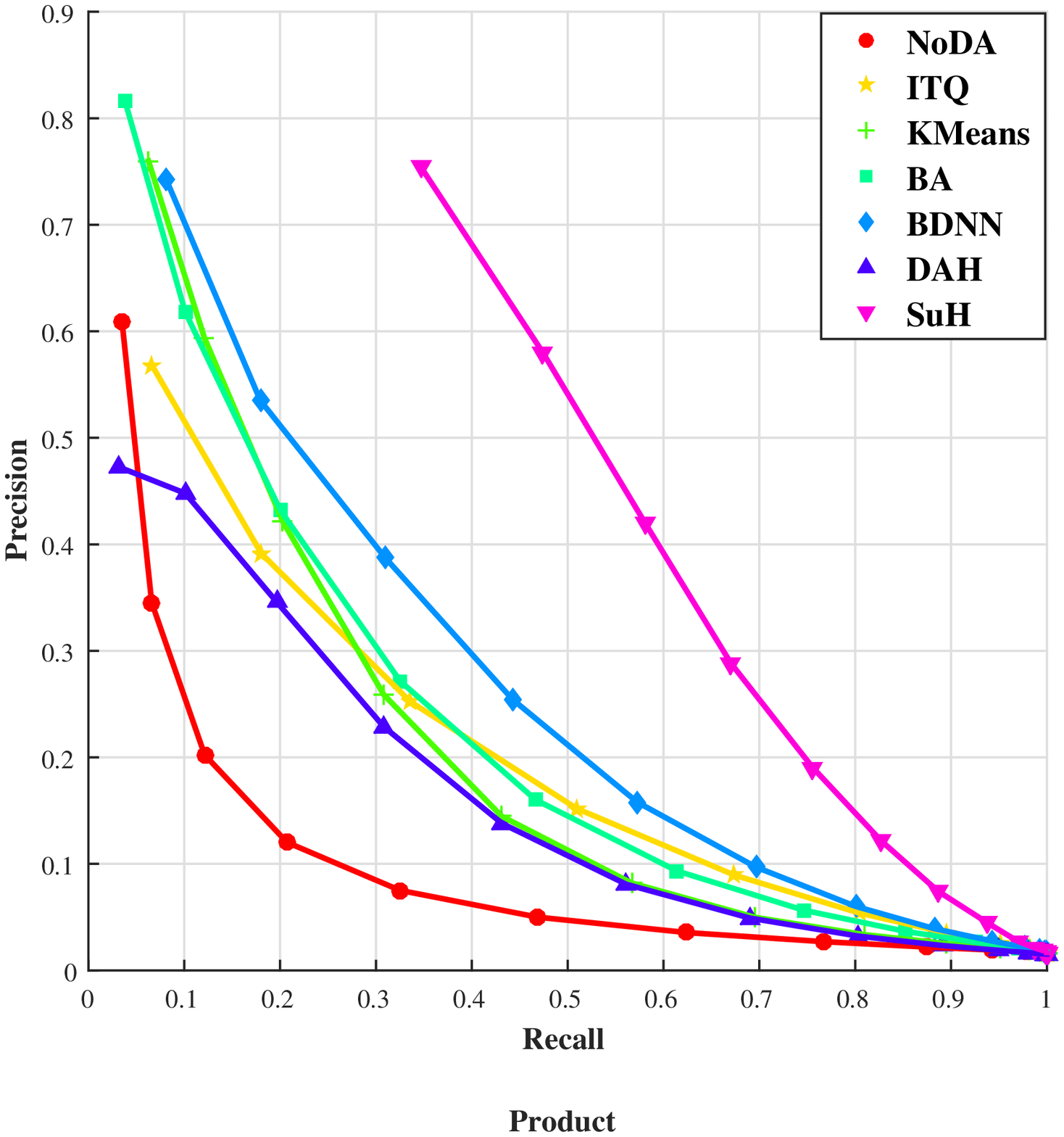}
}%
\subfloat[\scriptsize{Real-World}]{
		\label{Fig:RealWorldPR16}
    \includegraphics[trim = 35mm 14mm 38mm 10mm, clip, width=0.245\textwidth]{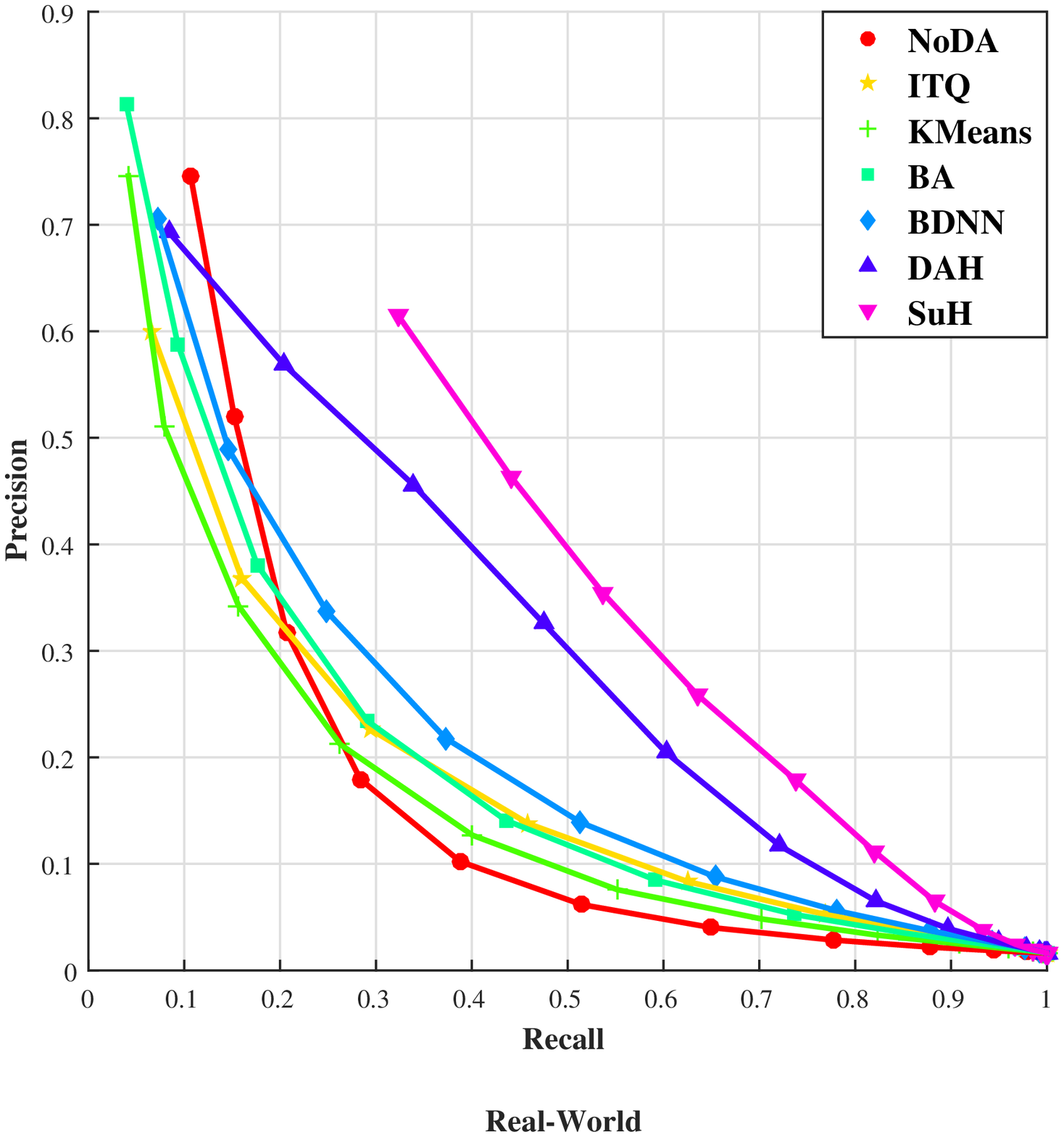}
}%
\caption{Precision-Recall curves @16 bits for the \textbf{\textit{Office-Home}} dataset. Comparison of hashing without domain adaptation (\textbf{NoDA}), shallow unsupervised hashing (\textbf{ITQ}, \textbf{KMeans}), state-of-the-art deep unsupervised hashing (\textbf{BA}, \textbf{BDNN}), unsupervised domain adaptive hashing (\textbf{DAH}) and supervised hashing (\textbf{SuH}). Best viewed in color.}
\label{Fig:OfficeHomePR16}
\end{figure*}

\begin{figure*}[!ht]
\centering
\subfloat[\scriptsize{Art}]{
		\label{Fig:ArtPR128}
    \includegraphics[trim = 35mm 14mm 41mm 10mm, clip, width=0.245\textwidth]{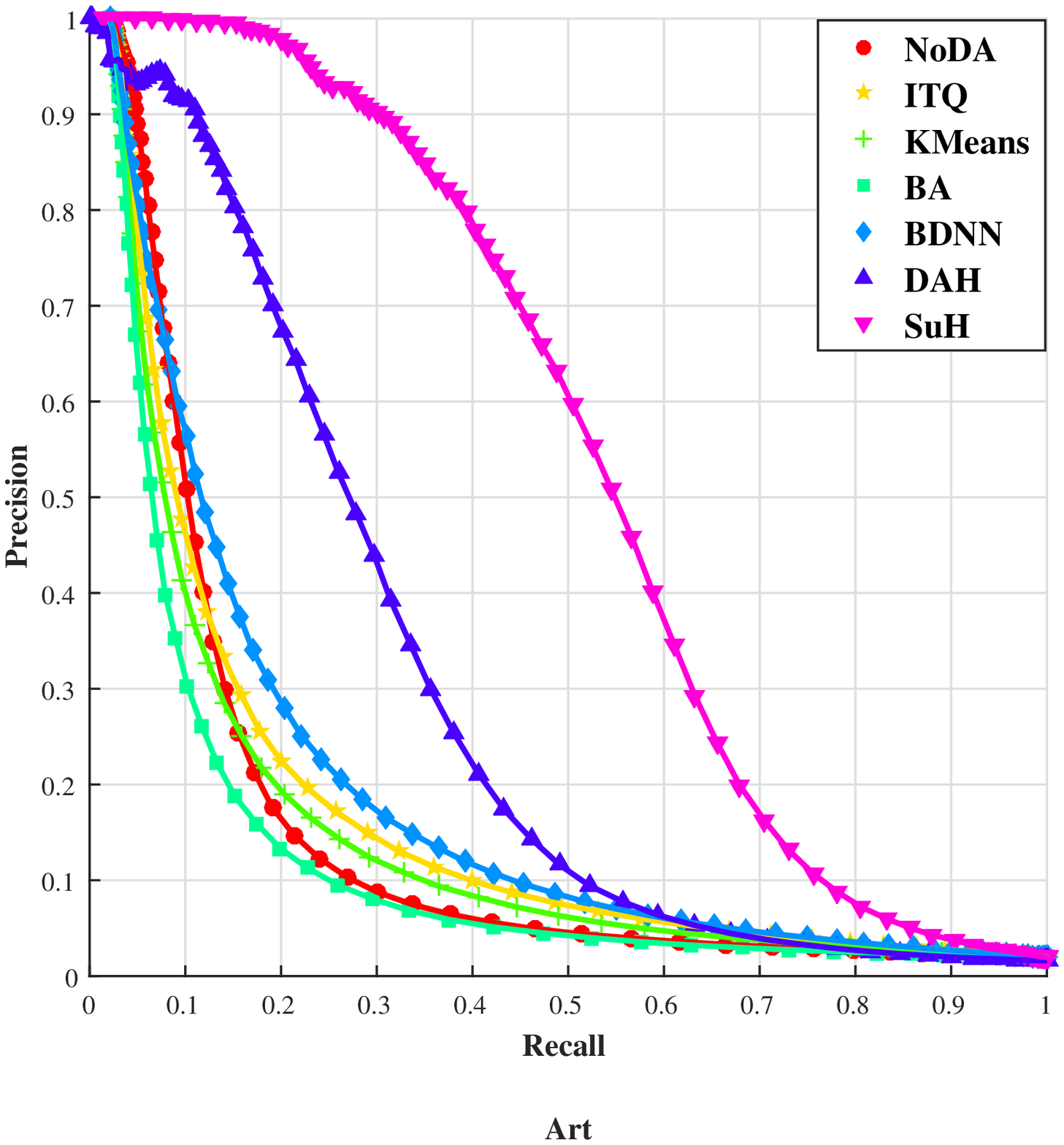}
}%
\subfloat[\scriptsize{Clipart}]{
		\label{Fig:ClipartPR128}
    \includegraphics[trim = 35mm 14mm 41mm 10mm, clip, width=0.245\textwidth]{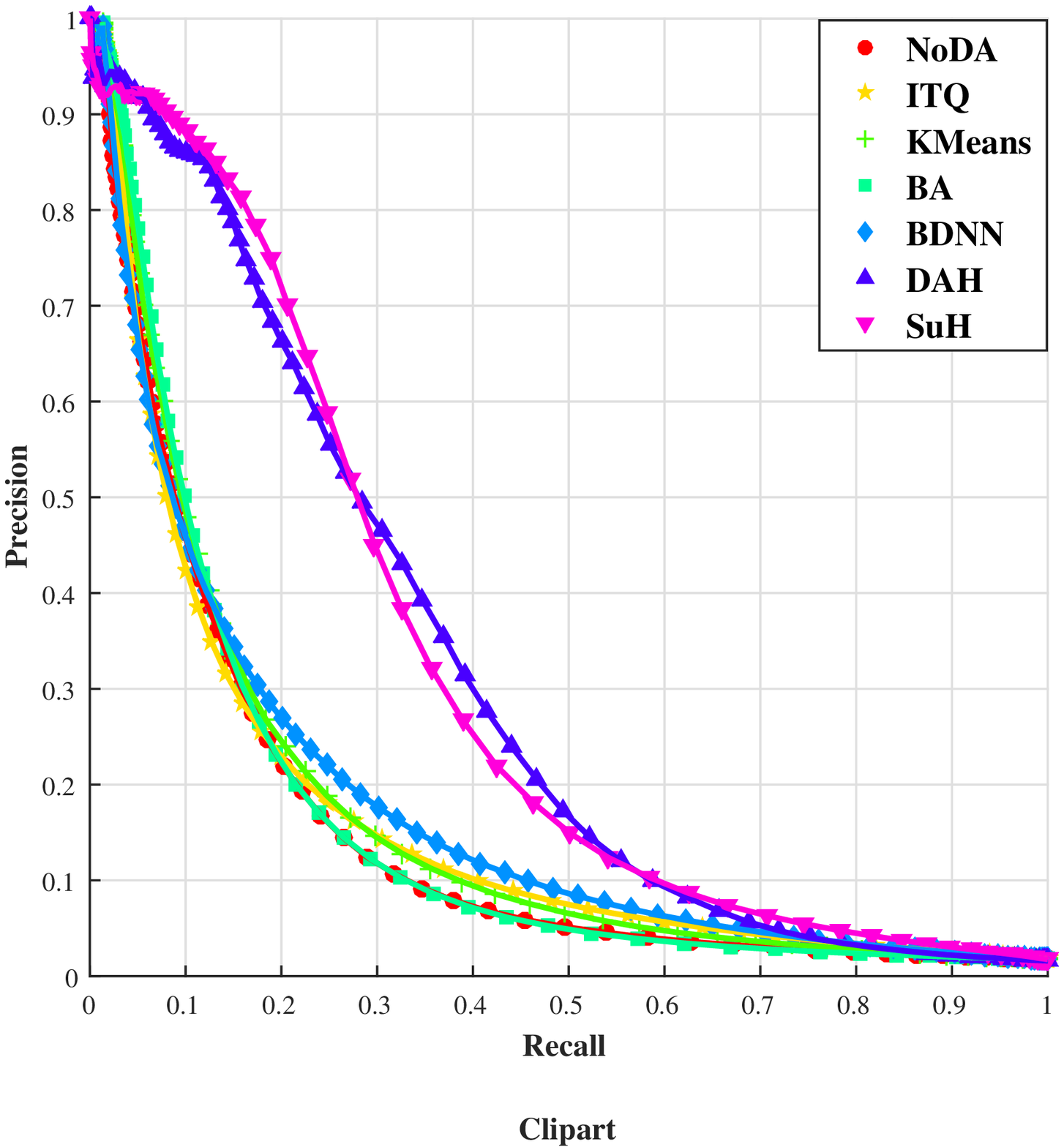}
}%
\subfloat[\scriptsize{Product}]{
		\label{Fig:ProductPR128}
    \includegraphics[trim = 35mm 14mm 41mm 10mm, clip, width=0.245\textwidth]{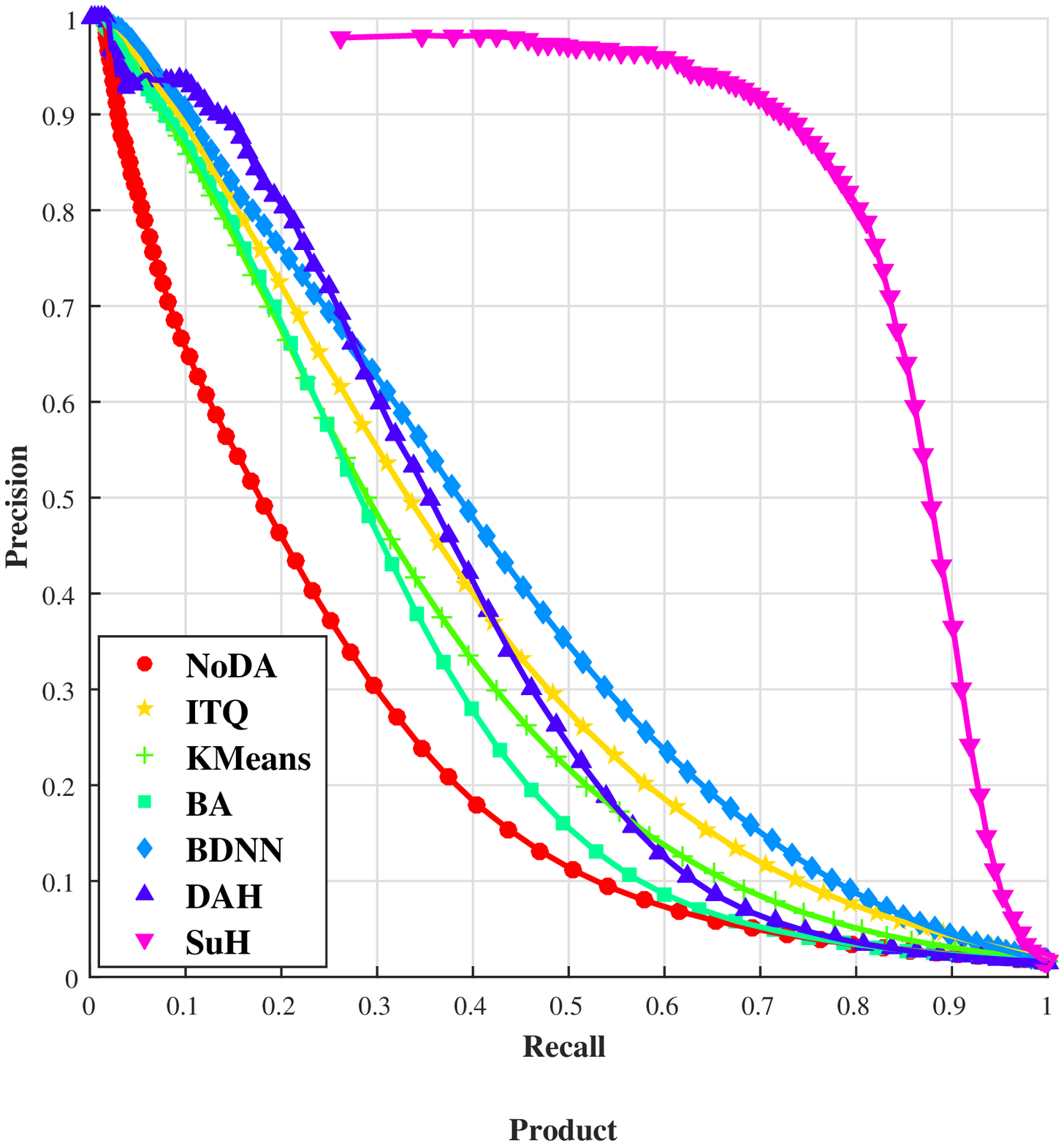}
}%
\subfloat[\scriptsize{Real-World}]{
		\label{Fig:RealWorldPR128}
    \includegraphics[trim = 35mm 14mm 41mm 10mm, clip, width=0.245\textwidth]{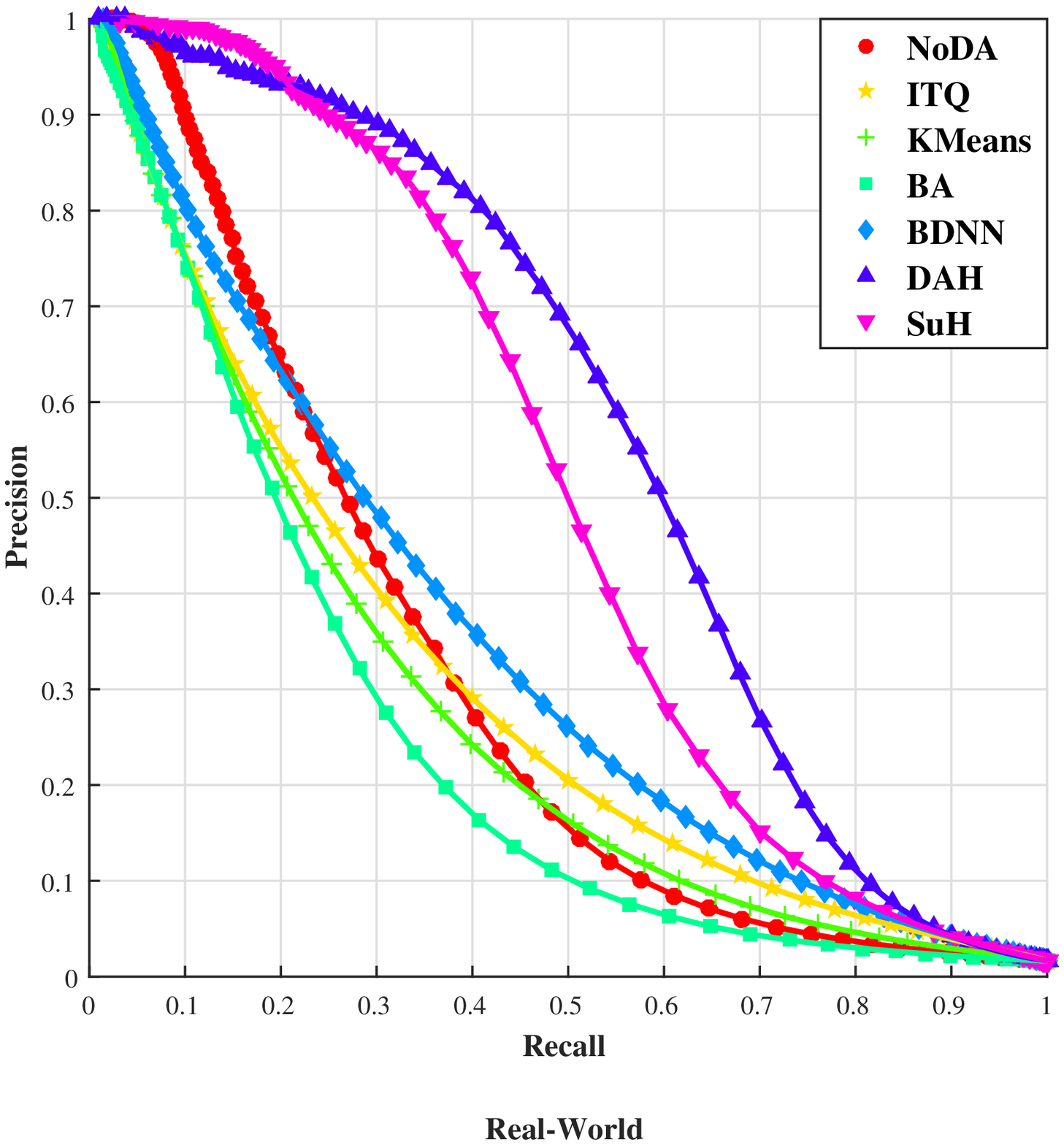}
}%
\caption{Precision-Recall curves @128 bits for the \textbf{\textit{Office-Home}} dataset. Comparison of hashing without domain adaptation (\textbf{NoDA}), shallow unsupervised hashing (\textbf{ITQ}, \textbf{KMeans}), state-of-the-art deep unsupervised hashing (\textbf{BA}, \textbf{BDNN}), unsupervised domain adaptive hashing (\textbf{DAH}) and supervised hashing (\textbf{SuH}). Best viewed in color.}
\label{Fig:OfficeHomePR128}
\end{figure*}

We provide the unsupervised domain adaptive hashing results for $d=16$ and $d=128$ bits in Figures \ref{Fig:OfficeHomePR16} and \ref{Fig:OfficeHomePR128} respectively. 
In Tables \ref{Tab:mAPHash16} and \ref{Tab:mAPHash128}, we outline the corresponding mAP values. The notations are along the lines outlined in the main paper. 
We observe similar trends for both $d=16$ and $d=128$ bits compared to $d=64$ bits. 
It is interesting to note that with increase in bit size $d$, the mAP does not necessarily increase. 
Table \ref{Tab:mAPHash128} ($d=64$) has its mAP values lower than those for $d=64$ (see main paper) for all the hashing methods. 
This indicates that merely increasing the hash code length does not always improve mAP scores. 
Also, the mAP values for \texttt{Real-World} for $d=128$ bits has DAH performing better than SuH. 
This indicates that in some cases domain adaptation helps in learning a better generalized model.  

\end{document}